\documentclass{article}

\usepackage{microtype}
\usepackage{graphicx}
\usepackage{subcaption}
\usepackage{booktabs} 

\usepackage{hyperref}

\usepackage[preprint]{icml2026}

\usepackage{amsmath}
\usepackage{amssymb}
\usepackage{mathtools}
\usepackage{amsthm}

\usepackage[capitalize,noabbrev]{cleveref}

\theoremstyle{plain}

\theoremstyle{definition}

\theoremstyle{remark}

\usepackage[textsize=tiny]{todonotes}

\icmltitlerunning{GRIP2: A Robust and Powerful Deep Knockoff Statistic for Feature Selection}

\begin{document}

\twocolumn[
  \icmltitle{GRIP2: A Robust and Powerful Deep Knockoff Method for Feature Selection}



  \icmlsetsymbol{equal}{*}

  \begin{icmlauthorlist}
    \icmlauthor{Bob Junyi Zou}{icme}
    \icmlauthor{Lu Tian}{dbds}
  \end{icmlauthorlist}

  \icmlaffiliation{icme}{Institute for Computational and Mathematical Engineering, Stanford University, Stanford, California, USA}
  \icmlaffiliation{dbds}{Department of Biomedical Data Science, Stanford University, Stanford, California, USA}

  \icmlcorrespondingauthor{Lu Tian}{lutian@stanford.edu}

  \icmlkeywords{Machine Learning, ICML}

  \vskip 0.3in
]



\printAffiliationsAndNotice{}  

\begin{abstract}
Identifying truly predictive covariates while strictly controlling false discoveries remains a fundamental challenge in nonlinear, highly correlated, and low signal-to-noise regimes, where deep learning based feature selection methods are most attractive. We propose \emph{Group Regularization Importance Persistence in 2 Dimensions (GRIP2)}, a deep knockoff feature importance statistic that integrates first-layer feature activity over a \emph{two-dimensional} regularization surface controlling both sparsity strength and sparsification geometry. To approximate this surface integral in a \emph{single} training run, we introduce efficient block-stochastic sampling, which aggregates feature activity magnitudes across diverse regularization regimes along the optimization trajectory. The resulting statistics are antisymmetric by construction, ensuring finite-sample FDR control. In extensive experiments on synthetic and semi-real data, GRIP2 demonstrates improved robustness to feature correlation and noise level: in high correlation and low signal-to-noise ratio regimes where standard deep learning based feature selectors may struggle, our method retains high power and stability. Finally, on real-world HIV drug resistance data, GRIP2 recovers known resistance-associated mutations with power better than established linear baselines, confirming its reliability in practice.
\end{abstract}

\section{Introduction}
\label{sec:intro}

Feature selection with rigorous false discovery rate (FDR) control is a core problem in scientific machine learning, with applications in genomics, medicine, and economics. In many such settings, the goal is not only accurate prediction but also reliable discovery: selected features are often by themselves important findings with downstream applications, so controlling the expected fraction of false discoveries is essential for reproducible scientific conclusions.

The Model-X knockoff framework provides a general route to controlled variable selection by constructing synthetic \emph{knockoff} variables that preserve the dependence structure of the covariates while remaining conditionally independent of the response \citep{barber2015controlling,candes2018panning}. Given any importance statistic computed on the augmented design consisting of both covariates and their knockoffs that is antisymmetric under swapping a feature with its knockoff, the knockoff filter guarantees finite-sample FDR control. This model-agnostic guarantee makes knockoffs especially attractive for extending FDR control beyond linear models, motivating a growing literature on nonlinear and deep knockoff procedures \citep{lu2018deeppink,zhu2021deeplink,zhu2021deep,oh2024deeppig}.

However, in the low signal-to-noise (SNR) and high correlation regimes that often motivate flexible and complex modeling, constructing stable and informative importance statistics remains challenging. Under strong correlation, many distinct feature subsets can yield essentially indistinguishable predictive performance, so attribution to individual variables becomes intrinsically ambiguous; this phenomenon is commonly referred to as the \emph{Rashomon} effect \citep{breiman2001statistical}. In this regime, post-hoc attribution scores~\citep{kassani2022deep,wang2025knockoff,mason2025novel} can vary substantially across equally accurate models~\citep{adebayo2018sanity,hooker2019benchmark}. Architecture-based deep knockoff methods such as DeepPINK~\citep{lu2018deeppink} explicitly encode feature--knockoff competition, but their importance depends on weights embedded in coupled nonlinear representations whose magnitudes may not reliably reflect feature relevance. 
More broadly, a critical limitation of existing nonlinear knockoff methods is their reliance on a fixed regularization geometry and a single hyperparameter setting.
In correlated designs, where several proxy features can explain the signal equally well, 
the specific choice of regularization strength determines which proxy is selected, introducing a source of arbitrariness into the discovery process.
Naive seed ensembling~\citep{hooker2019benchmark} can reduce variance from random initialization, but in correlated low-SNR settings it may offer limited gains because it does not resolve this structural instability and merely averages estimates produced under the same static regularization constraint (as we observe empirically) .

In this work, we propose \textbf{Group Regularization Importance Persistence in Two Dimensions (GRIP2)}, a deep knockoff importance statistic designed for variable selection in high correlation, low-SNR settings. GRIP2 imposes structured shrinkage on the first layer of a multi-layer perceptron (MLP) with the augmented inputs, where weights connected to each feature provide a direct and antisymmetry-compatible measure of that feature's activity. Instead of evaluating importance at a single regularization choice, GRIP2 aggregates aforementioned feature activity over a two-dimensional regularization surface that covers wide ranges of sparsity strength and sparsification geometry. Intuitively, true signal features tend to remain active across a broader range of regimes, while spurious or proxy features activate less consistently. To make this aggregation practical, we introduce block stochastic sampling, which traverses the regularization surface along a single training trajectory consisting of blocks of training steps and different sampled regularization parameters are used for updating within each block. This yields an efficient single-run approximation of the surface average, overcome the arbitrariness of fixed-point regularization.
The resulting statistic is antisymmetric by construction and therefore compatible with the Model-X knockoff filter, retaining finite-sample FDR guarantees.

Across synthetic experiments, semi-real experiments with signal injection, and a real HIV drug resistance data, we find that GRIP2 offers a favorable stability and power in the challenging high correlation, low-SNR settings.

\paragraph{Contributions.}
\begin{itemize}
    \item \textbf{Two-dimensional persistence for deep knockoff importance.}
    We propose a persistence-based importance statistic that aggregates first-layer feature activity over a regularization surface covering wide ranges of sparsity strength and sparsification geometry in deep learning to enhance robustness and power. 

    \item \textbf{Single-run estimation via block stochastic sampling.}
    We introduce a computationally efficient algorithm that approximates the surface-averaged importance statistic within a single training run, exploring multiple regularization regimes and mimicking seed ensembles.

    \item \textbf{Empirical study in high correlation, low-SNR regimes.}
    We evaluate GRIP2 against common nonlinear and linear baseline methods with matched computation budgets, and demonstrate superior stability, power and FDR control.

    \item \textbf{Real-world validation.}
    We apply GRIP2 to recover drug-resistance mutations by analyzing a real HIV drug data and demonstrate a better performance than established linear baselines in the literature.
\end{itemize}



\section{Methodology}
\label{sec:method}

\subsection{Problem Setup and Model-X Framework}
\label{subsec:setup}

We consider the supervised learning task with i.i.d.\ samples $\{(x_i, y_i)\}_{i=1}^n$ drawn from a joint distribution $P_{XY}$, where $X \in \mathbb{R}^p$ and $Y \in \mathbb{R}$ is the response. Our goal is to identify the set of active features $\mathcal{S} = \{j : X_j \not\perp Y \mid X_{-j}\}$ while controlling the False Discovery Rate (FDR)--the expected proportion of irrelevant features among those selected.

To this end, we employ the \emph{model-X knockoff} framework from \citet{candes2018panning}. For a given $X$, we construct knockoff variables $\tilde X \in \mathbb{R}^p$ satisfying the \emph{swap invariance} property: for any subset of indices $G \subset \{1, \dots, p\}$,
\begin{equation}
\label{eq:exchangeability}
    (X, \tilde X)_{\text{swap}(G)} \stackrel{d}{=} (X, \tilde X).
\end{equation}
By augmenting the feature space to $(X, \tilde X) \in \mathbb{R}^{2p}$, we create a controlled experiment where $\tilde X_j$ serves as a negative control for each $X_j$. Feature selection can be conducted based on a set of importance statistics, $W_1, \cdots, W_p,$ derived using augmented features. To achieve the FDR control, the importance statistic must satisfy the \emph{antisymmetry property}: swapping $X_j$ and $\tilde X_j$ must flip the sign of $W_j$.

\subsection{Neural Model with First-layer Group Regularization}
\label{subsec:model}
We parameterize the predictive model as a feedforward neural network $f_\theta$ taking augmented input $X^{\text{aug}}=(X,\tilde X)\in\mathbb{R}^{2p}$.   
Let $W=[w_1, \cdots, w_{2p}] \in \mathbb{R}^{d \times 2p}$ denote the first-layer weight matrix, where  $w_j\in\mathbb{R}^d$ is its $j$th column corresponding to feature $j$ (original or knockoff), acting as a ``gate'' for feature $j$ entering the neural network. 
We introduce a group regularization for $w_j$ in training the neural network and 
parameterize regularization by a pair $(\lambda,a)$. Specifically, the loss function ${\cal L}(\theta; \lambda, a)$ is 
\begin{equation}
\label{eq:objective}
\begin{split}
   \frac{1}{n}\sum_{i=1}^n \ell(y_i, f_\theta(x_i^{\text{aug}})) 
   + \lambda \sum_{j=1}^{2p} \|w_j\|_2^a + \frac{\gamma}{2} \|\theta_{\text{deep}}\|_2^2,
\end{split}
\end{equation}
\begin{itemize}
\item where the $\ell_2$ norm $\|w_j\|_2$ represents the ``activity'' of feature $j\in \{1, \cdots, 2p\},$
\item $\theta_{\text{deep}} = \theta \setminus \{W\}$ represents all MLP parameters beyond the first layer, on which the standard $l_2$ regularization is still applied to ensure the identifiability,
\item $\lambda>0$ controls the overall sparsity strength,
\item and $a\in(0,1]$ controls the \emph{geometry} of group penalty.
\end{itemize}

\textbf{Remark 1.} Due to the scaling symmetry of neural networks, the model may  minimize the group penalty $\sum_{j=1}^{2p} \|w_j\|_2^a$ by shrinking $W$ toward zero, while scaling up the corresponding components of $\theta_{\text{deep}}$ without changing the output of the MLP. We therefore apply $l_2$ regularization to $\theta_{\text{deep}}$ to prevent this phenomenon so that $\|w_j\|_2$ remains as a faithful proxy for the feature activity in the neural network.

\textbf{Remark 2.} Our intention is to use $\|w_j\|_2$, the feature ``activity", as a importance measure. While it appears naive to summarize feature importance only using the first layer weights of a MLP, this method is equivalent to using $|c_j|$ to measure feature importance with the following loss function in training the neural network:
$$
   \frac{1}{n}\sum_{i=1}^n \ell(y_i, f_\theta(c \odot x_i^{\text{aug}})) + \lambda_a\sum_{j=1}^{2p} |c_j|^b + \frac{\gamma_\theta}{2} \|\theta\|_2^2,
$$
where $c=(c_1, \cdots, c_{2p})',$ $b$ is a  positive constant, and $\odot$ stands for point-wise product. Clearly, $|c_i|$ controls the ``role'' of feature in the MLP by rescaling it as an input and thus is a natural measure for feature importance. 


\subsection{Group Activity and the Regularization Surface}
\label{subsec:activity}

For a given pair $(\lambda, a)$, we define the \emph{group activity} $A_j(\lambda, a)$ as the $\ell_2$ norm of the first-layer weights at a local minimizer of \eqref{eq:objective}, 
$\theta^*(\lambda, a) = \{W^*(\lambda, a), \theta_{\text{deep}}^*(\lambda, a)\}:$ 
\begin{equation}
\label{eq:activity_def}
A_j(\lambda, a) = \|w_j^*(\lambda, a)\|_2.
\end{equation}
This group activity is a scalar, permutation-equivariant quantity that directly reflects whether feature $j$ contributes to the learned representation. In this non-convex setting, $A_j(\lambda, a)$ characterizes the ``contribution" of feature $j$ to an equilibrium state, where the gradient of the predictive loss is balanced by that from the sparsity-inducing penalty in (\ref{eq:objective}). By varying $(\lambda, a)$, we explore the feature importance over a 2D \emph{regularization surface}.

\subsection{Two-Dimensional Soft Regularization Persistence}
\label{subsec:persistence}

We propose a \emph{soft regularization persistence}, measuring feature importance over the $(\lambda, a)$ surface. Specifically, let $\mu$ be a ``prior'' distribution for $(\lambda, a)$ over $\Omega = [\lambda_{\min}, \lambda_{\max}] \times [a_{\min}, 1],$ where $\lambda_{\min}, \lambda_{\max}$ and $a_{\min}$ are suitable constants. We define the \emph{soft regularization persistence score} for the $j$th feature as:
\begin{equation}
\label{eq:persistence}
S_j = \mathbb{E}_{(\lambda, a) \sim \mu} \left[ A_j(\lambda, a) \right] = \int_\Omega A_j(\lambda, a) \, \mu(d\lambda, da),
\end{equation}
where $(\log (\lambda), a) \sim \mathcal{U}\left([\log (\lambda_{\min}), \log (\lambda_{\max})]\times [a_{\min}, 1]\right).$

\textbf{Intuition on Sparsity Geometry.} The parameter $a$ controls the concavity of the penalty. While $a=1$ (Group Lasso) is convex, it induces heavy shrinkage bias. Lowering $a$ moves towards non-convex regularization that approximate $L_0$-based feature selection, forcing ``competitive exclusion'' among correlated features with less bias. A signal is \emph{persistent} if the corresponding $w_j$ survives both global pressure $\lambda$ and the aggressive selection geometry of $a < 1$. Intuitively, informative features corresponding to true signals are expected to remain active across a wide range of sparsification regimes, whereas null features activate, i.e., with a large $A_j(\lambda, a),$ only sporadically at the most. Therefore, the soft regularization persistence scores of predictive features tend to be greater than those of null features. 

\textbf{Technical Detail: Calibrating $\Omega$ via Gradient Ratios.}
To automate the adaptive selection of the support $[\lambda_{\min}, \lambda_{\max}]\times[a_{\min},1]$, we perform a short calibration run to balance the regularization strength against the predictive signal. This dynamic calibration removes the need for manual grid search by anchoring the regularization strength to the empirical signal level. We select a support such that the gradient magnitudes of the regularization penalty $\mathcal{R}(\theta) = \lambda \sum \|w_j\|_2^a$ and that of the predictive loss $\mathcal{L}_{\text{pred}}(\theta)=n^{-1}\sum_{i=1}^n \ell(y_i, f_\theta(x_i^{\text{aug}})) $ are comparable:
\begin{equation}
    \frac{\|\nabla_W \mathcal{R}(\theta)\|_F}{\|\nabla_W \mathcal{L}_{\text{pred}}(\theta)\|_F} \;\in\; [r_{\min}, r_{\max}].
\end{equation}
This criterion ensures that the regularization force is sufficiently strong to induce sparsity  without overwhelming gradient information from training data. For simulation settings where the covariance matrix is well-conditioned and knockoff construction is exact, we recommend $[r_{\min}, r_{\max}]=[0.01, 0.20]$. When the covariance matrix is ill-conditioned and knockoffs are approximate, we recommend a wider range $[r_{\min}, r_{\max}]=[0.01, 1].$ Lastly, to encourage the exploration of the regularization geometry, we set $a_{\min}=0.1.$  


\subsection{Efficient Approximation via BSS}
\label{subsec:block}

Computing the proposed score $S_j=\mathbb{E}_{(\lambda,a)\sim\mu}[A_j(\lambda,a)]$ by training a dense grid over $(\lambda,a)$ can be expensive. We therefore introduce \emph{Block Stochastic Sampling} (BSS) algorithm to estimate $S_j$ within a \emph{single} training run traversing of the entire regularization surface. Algorithm~\ref{alg:bss} summarizes the procedure: we partition the training run into $B$ blocks of $M$ minibatch updates; at the start of block $b$, we sample $(\lambda_b,a_b)\sim\mu$ and optimize the regularized objective for $M$ steps; at the end of the block, we record the first-layer group magnitudes
$a_j^{(b)}=\|w_j\|_2$ for $j=1,\dots,2p$, and return $\widehat{S}_j \;=\; B^{-1}\sum_{b=1}^B a_j^{(b)}.$

\textbf{Remark 3.}
BSS can be viewed as a \emph{mixture-of-regimes} estimator: each block produces a snapshot of the gate magnitudes under a particular sparsification regime $(\lambda_b,a_b)$. If block is large enough for the optimizer to converge under the current regularization, then $a_j^{(b)}$ is a good proxy for the equilibrium activity $A_j(\lambda_b,a_b)$ and $\widehat{S}_j$ approximates the target surface average $S_j$.

\textbf{Block calibration.}
The primary computational parameter of the algorithm is the block size $M$, which should exceed the number of updates required for first-layer group norms to respond to changes in $(\lambda,a)$. In practice, to avoid excessive computation with an overly large $M$, we calibrate it using a small pilot run by monitoring the average absolute change of group norms,
\begin{equation}
\Delta(t)
~:=~
\frac{1}{2p}\sum_{j=1}^{2p}
\big|\|w_j^{(t)}\|_2-\|w_j^{(t-1)}\|_2\big|,
\end{equation}
where $w_j^{(t)}$ is the value of $w_j$ at the $t$th update within a block. 
We choose $M$ such that $\Delta(t)$ 
remains below a threshold $\delta$ 
for a sustained length of updates across all blocks. This criterion ensures that the recorded snapshot $a_j^{(b)}$ reflects the current regime rather than transient optimization lag immediately after switching $(\lambda,a)$. 
Since we only require the stabilization of the \emph{first-layer} weights $W$, not full convergence of all network parameters, moderate block sizes are oftentimes sufficient in practice. In our experiments, we use $M=25$ for synthetic/semi-real settings, and $M=50$ for real settings respectively.  

\begin{algorithm}[h!]
\caption{BSS for Persistence Scores}
\label{alg:bss}
\begin{algorithmic}[1]
\STATE \textbf{Input:} Training data $\{x_i, \tilde x_i, y_i\}$, distribution $\mu$, block size $M$, iterations $T$, weight decay $\gamma$.
\STATE \textbf{Initialize:} Model parameters $\theta_0 = \{W_0, \theta_{\text{deep}, 0}\}$.
\FOR{block $b = 1, \dots, B=\lfloor T/M \rfloor$}
    \STATE Sample $(\lambda_b, a_b) \sim \mu$.
    \FOR{step $t = 1, \dots, M$}
        \STATE Sample minibatch $\mathcal{B}$.
        \STATE Compute gradient $\nabla_\theta \mathcal{L}(\theta; \lambda_b, a_b)$ using Eq.~\eqref{eq:objective}.
        \STATE Update $\theta \leftarrow \theta - \eta \nabla_\theta \mathcal{L}$.
    \ENDFOR
    \STATE Record feature activity: $a_j^{(b)} \leftarrow \|w_j\|_2,$ $j=1, \dots, 2p$.
\ENDFOR
\STATE \textbf{Return:} Estimated persistence $\widehat S_j = B^{-1} \sum_{b=1}^B a_j^{(b)}$.
\end{algorithmic}
\end{algorithm}

\subsection{Knockoff Statistic and Antisymmetry}
\label{subsec:antisymmetry}

For each original feature $j \in \{1, \dots, p\}$, we define the importance statistic $W_j = S_j - S_{j+p}$. 

\textbf{Proposition 1 (Antisymmetry).} 
The statistic $W_j$ satisfies the antisymmetric property:
\begin{equation}
    W_j((X, \tilde X)_{\text{swap}(G)}, Y) = 
    \begin{cases}
    -W_j((X, \tilde{X}), Y),\,\, &j\in G,\\
    W_j((X, \tilde{X}), Y),\,\, &j\notin G,\\
    \end{cases}
\end{equation}
where $G$ is any set of feature indices and $(X, \tilde{X})_{\text{swap}(G)}$ is the feature matrix obtained by swapping columns in $G$.

\textbf{Proof.} 
Let $Z = (X, \tilde{X})$ be the augmented design matrix. Consider the swap operation for a fixed feature $j$, denoted by $Z_{{\cal S}_j} = (X, \tilde{X})_{\text{swap}(\{j\})}$, which exchanges columns $j$ and $j+p$.
The objective function \eqref{eq:objective} depends on the data solely through the terms $f_\theta(x_i^{\text{aug}})$. 
Due to the structure of the first layer, swapping the $j$-th and $(j+p)$-th input features is functionally equivalent to swapping the corresponding weight vectors $w_j$ and $w_{j+p}$.
Consequently, the objective function satisfies the symmetry property $\mathcal{L}(\{w_j, w_{j+p}, \dots\} \mid Z_{{\cal S}_j}, Y) = \mathcal{L}(\{w_{j+p}, w_j, \dots\}\mid Z, Y)$.
Since the regularization penalty and the distribution $\mu$ are symmetric with respect to indices $j$ and $j+p$, the minimizer (or equilibrium path) $\theta^*$ is equivariant under this permutation.
Therefore, the persistence scores satisfy $S_j(Z_{{\cal S}_j}, Y) = S_{j+p}(Z, Y)$ and $S_{j+p}(Z_{{\cal S}_j}, Y) = S_j(Z, Y)$.
Substituting these into the definition of $W_j$, we obtain
$ W_j(Z_{{\cal S}_j}, Y) = S_j(Z_{{\cal S}_j}, Y) - S_{j+p}(Z_{{\cal S}_j}, Y)= S_{j+p}(Z, Y) - S_j(Z, Y) = -W_j(Z, Y).$  For any $k \notin \{j, j+p\}$, the inputs and weights remain unchanged, implying $W_k(Z_{{\cal S}_j}, Y) = W_k(Z, Y)$. The same argument can be used to show (7) holds for general $G.$
\hfill $\square$

\textbf{Remark 4.} When knockoff exchangeability (Equation~\ref{eq:exchangeability}) is satisfied, this Proposition ensures GRIP2 has valid FDR control in finite sample~\citep{candes2018panning}.

\section{Related Work}
\label{sec:related_work}

\textbf{Architecture-based deep knockoff statistics.}
Several feature selection methods build feature--knockoff competition into the network architecture. DeepPINK~\citep{lu2018deeppink} introduces a pairwise-connected layer that couples each $(X_j,\tilde X_j)$, and subsequent work extends this idea using learned knockoff generators and gating variants, e.g., DeepLINK~\citep{zhu2021deeplink}, Deep-gKnock~\citep{zhu2021deep}, and DeepPIG~\citep{oh2024deeppig}. These approaches 
typically rely on a \emph{fixed} regularization configuration and in the highly correlated and low-SNR regimes that motivate our work, 
a fixed regularization configuration may force an arbitrary selection among correlated features.
Moreover, many architecture-driven statistics ultimately depend on weights embedded in coupled nonlinear representations, where magnitudes can be difficult to interpret as feature importance.

\textbf{Post-hoc importance with knockoffs.}
Another line of work separates prediction from selection by 
constructing 
antisymmetric knockoff statistics based on a trained predictor. Examples include 
knockoffs with SHAP-type scores,  gradient-based sensitivity statistics and other feature importance measures
~\citep{kassani2022deep,wang2025knockoff,mason2025novel}. A key limitation in high correlation and low-SNR settings is that post-hoc attributions are tied to a \emph{single} snapshot of the optimization trajectory: in the presence of the Rashomon effect (i.e. when multiple models yield similar predictions), relying on one specific solution introduces structural instability, as the importance scores may reflect initialization noise or a specific local minimum rather than robust statistical relevance, and thus can vary across those equally predictive models. 

\textbf{Path-based ranking and regularization.}
Recent work shows that leveraging information along a regularization path (rather than a single solution) can better separate signals from noise features in linear models~\citep{weinstein2023power,ke2024power}. Our method extends this perspective to nonlinear model with flexible nonconvex regularization, and define importance measures via persistence across regimes over a regularization surface.
By integrating over this surface, our method mitigates the instability of point-estimation approaches, generalizing the benefits of path-based ranking to nonconvex deep learning settings. 

\textbf{Multiple Knockoffs.}
Another strategy to stabilize selection is to generate multiple independent knockoff copies for each feature \citep{gimenez2019improving,he2021identification} or to aggregate results across multiple random knockoff realizations \citep{ren2023derandomizing, ren2024derandomised}, which increases the effective sample size of the null distribution and improves power. While effective, this approach requires constructing multiple valid knockoff copies, which can be expensive or challenging when the covariate distribution is high-dimensional and complex. Our method is complementary to the multiple-knockoff strategy: GRIP2 can stabilize importance estimates for each given set of knockoffs. 

\section{Experiments}
\label{sec:experiments}

We evaluate whether the proposed \emph{2D soft regularization-persistence} statistic improves variable-selection accuracy and robustness, with a particular focus on high correlation and low SNR regimes. Selected evaluation metrics and baseline methods are summarized in Table~~\ref{tab:baselines}. Specific implementation details are reported in Appendix~\ref{app:impl}.

\begin{table}[t!]
\centering
\small
\begin{tabular}{p{2.6cm} p{5.2cm}}
\hline
\multicolumn{2}{c}{\textbf{Metrics}} \\[0.5em]
\textbf{False Discovery Rate (FDR)} &
Expected proportion of null features among the selected set. \\[0.5em]

\textbf{Power} &
Expected proportion of true signals correctly identified. \\[0.5em]

\textbf{Selection Stability} &
Average pairwise Jaccard similarity of selected feature sets across independent trials with fixed ground-truth support but varying noise realizations and model initializations, measuring robustness to optimization stochasticity. \\[0.5em]


\multicolumn{2}{c}{\textbf{Deep Feature Selection Baselines}} \\[0.5em]
\textbf{DeepPINK (DP)} &
\citet{lu2018deeppink}: Neural knockoff filter using architecture-level competition between original features and their knockoffs. \\[0.5em]

\textbf{MALD} &
\citet{mason2025novel}: Gradient-based sensitivity statistic applied to knockoff-augmented neural networks. \\[0.5em]

\textbf{SHAP} &
\citet{shrikumar2017learning}; \citet{wang2025knockoff}: Post-hoc attribution methods applied to a trained model, combined with Model-X knockoffs for feature selection. \\[0.5em]

\textbf{Lasso Path Entry Time (LAPA)} &
\citet{barber2015controlling}: Linear knockoff statistic based on feature entry time along the Lasso regularization path. \\[0.5em]

\textbf{Seed Ensembling} &
 Averaging importance scores across multiple random seeds to evaluate the performance for all above methods. \\[0.5em]

\multicolumn{2}{c}{\textbf{Ablation Variants}} \\[0.5em]
\textbf{1D Path ($\lambda$) (GRIP1)} &
Integration over regularization strength $\lambda$ with fixed sparsity geometry ($a=1$). \\[0.5em]

\textbf{1D Path ($a$) (GRIP1a)} &
Integration over sparsity geometry $a$ with fixed $\lambda=\sqrt{\lambda_{\min}\lambda_{\max}}$. \\[0.5em]

\textbf{Single-shot Group LASSO (GR)} &
Fixed regularization with $\lambda=\sqrt{\lambda_{\min}\lambda_{\max}}$ and $a=1$. \\[0.5em]

\multicolumn{2}{c}{\textbf{Computational Cost}} \\[0.5em]
\textbf{Single-run} &
 For all neural network based model, we use the same number of hidden layers and units and train for 5000 mini-batch steps on a single GPU in each trial. For Lasso path, we set the maximum number of iterations to 5000. \\[0.5em]
 
\textbf{K-level Seed Ensemble} &
 Repeat each trial with $K$ different seeds, leading to a computational cost about $K$ times that of a single-run.\\[0.5em]

\hline
\end{tabular}
\caption{Metrics, baselines, ablations, and computational cost used across all experiments.}
\label{tab:baselines}
\end{table}

\subsection{Synthetic Data Experiments}
\label{subsec:synthetic}

\textbf{Data Generation.} We simulate a high-dimensional regression task with $n=20000$, $p=500$, where covariates $X \sim \mathcal{N}(0,\Sigma)$ are generated from an adversarial first-order auto-regressive model designed to confuse feature selectors. The response is given by a nonlinear single-index model: 
\begin{equation} 
\begin{split} 
&Y=\sin\left(\frac{X^T\beta}{\sqrt{|S|}}\right)+\varepsilon, \quad \varepsilon\sim \mathcal{N}(0,\sigma^2)\\ 
&\beta_S\sim \mathcal{N}(0,I),\quad \beta_{-S}=0, 
\end{split} 
\end{equation} 
We focus on an \textbf{adversarially correlated} regime: 
\begin{itemize} 
\item \textbf{Correlation:} $\Sigma_{ij} = \rho^{|i-j|}$ with $\rho \in [0, 0.8] $.
\item \textbf{Support Structure:} $S = \{5k \mid k \in \mathbb{Z}_{100}\}$. By spacing signals at regular intervals, we ensure every true signal $X_j$ is flanked by highly correlated nulls ($X_{j-1}, X_{j+1}$) that act as ``decoys." 
\item \textbf{SNR:} The noise variance $\sigma^2$ is calibrated such that $\text{Var}(E(Y|X))/\sigma^2 = 0.2$ (Low SNR).
\end{itemize} 
To measure stability, we fix the support $S$ and coefficient vector $\beta$ across trials, while resampling $X$ and $\varepsilon$. In addition, the knockoffs are generated using the standard second-order Gaussian knockoff procedure described in \citet{candes2018panning},  details of which are provided in Appendix~\ref{app:gaussian_knockoffs}. 

\textbf{Results and Discussion.}

\textbf{FDR Control.} All methods exhibit valid FDR control across correlation settings. Since we use exact Gaussian knockoff construction and antisymmetric importance statistics, FDR control is theoretically guaranteed and empirically satisfied. We therefore 
only report them in Appendix~\ref{app:fdr_plots} (Figure~\ref{fig:app_fdr_control}).

\begin{figure}[t!]
    \centering
    \includegraphics[width=0.80\linewidth]{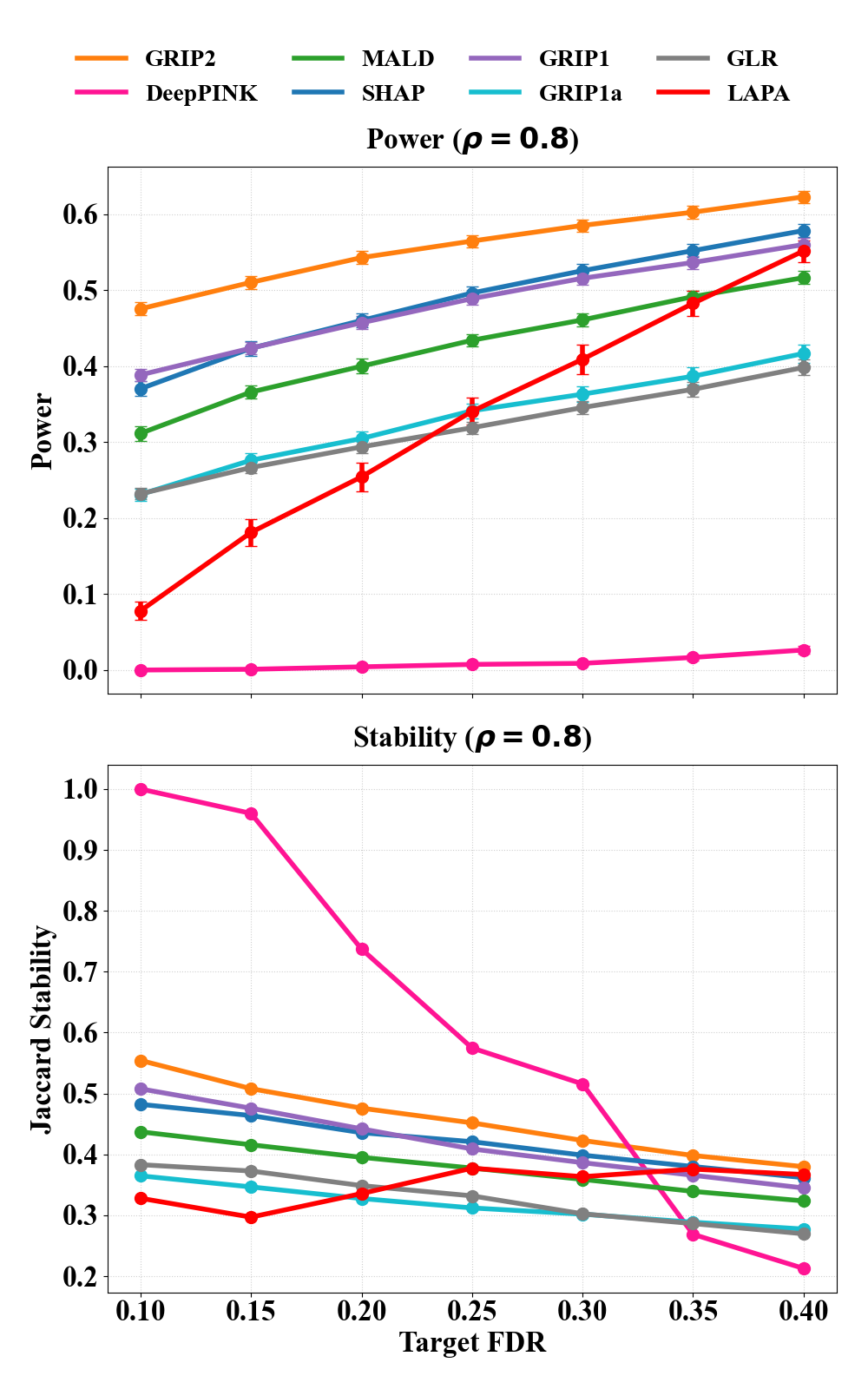}
    \caption{\textbf{Adversarial high-correlation regime ($\rho=0.8$).}
Power (top) and selection stability (bottom; average pairwise Jaccard) versus target FDR $q$. Power increases with $q$ while stability typically declines. Compared to other methods, GRIP2 maintains a favorable power--stability trade-off across $q$.}
\label{fig:synth_rho08}

\end{figure}

\textbf{Power and Stability under High Correlation.}
We focus on the challenging setting $\rho=0.8$, where each true signal competes with highly correlated decoy features.
Figure~\ref{fig:synth_rho08} illustrates the expected monotone trend of power function: power rises with the targeted FDR for all methods, while stability generally declines.
Across the full range of target FDR values, GRIP2 consistently achieves higher power and better stability while avoiding the abrupt stability deterioration observed in DeepPINK.

\begin{figure}[t!]
    \centering
    \includegraphics[width=0.80\linewidth]{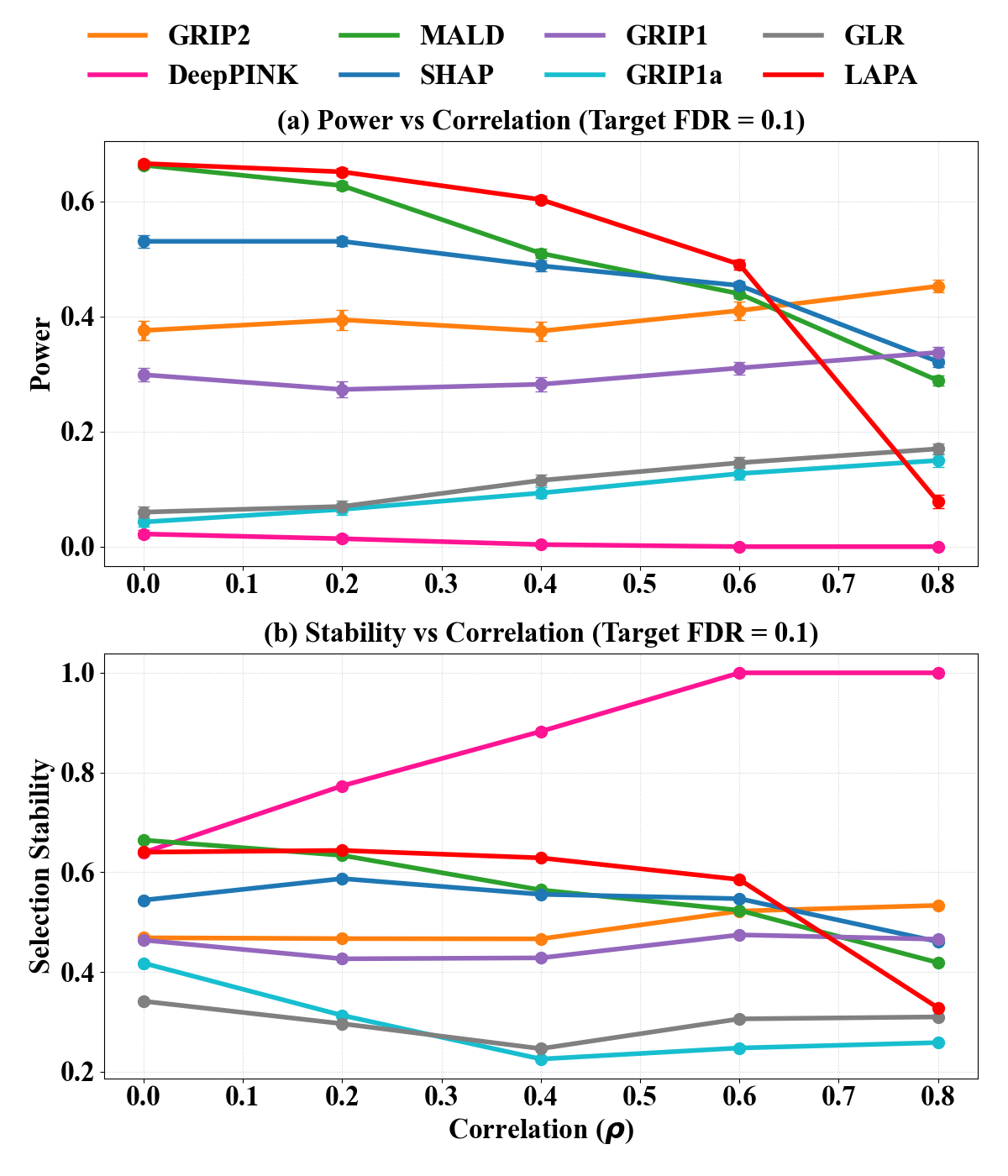}
    \caption{\textbf{Robustness to correlation strength.}
    Power (a) and stability (b) as a function of $\rho \in [0,0.8]$ at target FDR $q=0.1$.
    Competing methods exhibit a degradation as correlation increases, whereas GRIP2 performs better at higher $\rho$, indicating robustness to adversarial correlation through integration over sparsity geometries.}
    \label{fig:synth_rho_sweep}
\end{figure}

\textbf{Sensitivity to Correlation Strength.}
To examine the effect of correlation strength, Figure~\ref{fig:synth_rho_sweep} sweeps $\rho$ while fixing the target FDR at $q=0.1$.
While non-GRIP baseline methods exhibit power and stability deterioration as $\rho$ increases with a pronounced degradation as $\rho$ increases above 0.6, GRIP2's performance is stable and notably better in high correlation regimes, achieving the best power and stability at $\rho=0.8$, consistent with the hypothesis that varying both regularization strength and geometry helps break ties among correlated predictors. 

\begin{figure}[t!]
    \centering
    \includegraphics[width=0.80\linewidth]{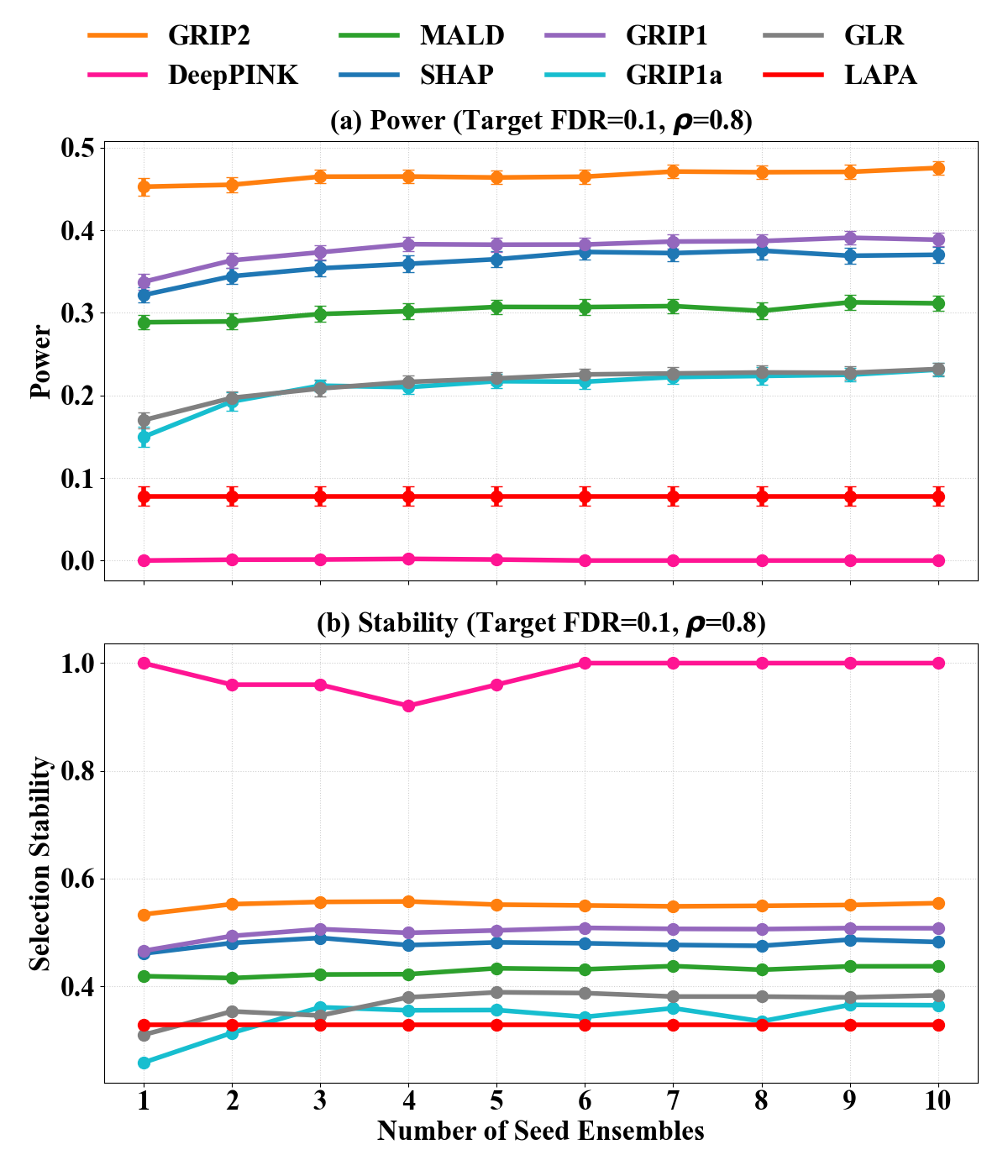}
    \caption{\textbf{Effect of seed ensembling ($\rho=0.8$, target FDR $q=0.1$).}
    Power (a) and stability (b) versus ensemble size $K$.
    Increasing $K$ yields limited gains for most baselines and does not resolve structural instability caused by correlated decoys.
    GRIP2 achieves strong stability and power with a single training run without requiring seed ensembling} 
    \label{fig:synth_ensembles}
\end{figure}

\textbf{Efficiency and the Role of Ensembling.}
Figure~\ref{fig:synth_ensembles} evaluates the stability/power vs computational cost trade-off by implementing seed ensembles of different methods in the comparison. In our setting, the effect of seed ensembles is limited and inconsistent, suggesting that simply averaging across random initializations does not reliably address ambiguity induced by strong feature correlation.
Among competitive methods (excluding DeepPINK as its extremely low power renders high stability meaningless), GRIP2 attains the best power in a single training run without requiring seed ensembling, indicating that its superiority arises from exploring the surface of regularization regimes rather than from averaging different random initializations.

\subsection{Semi-Real Experiments with Signal Injection}
\label{subsec:semireal}

\textbf{Dataset and Setup.}
To evaluate performance on a more realistic feature distribution with known ground truth, we adopt a semi-real protocol: covariates $X$ are drawn from real data, while responses $Y$ are synthetically generated with a fixed sparse support. 

We use the Human Activity Recognition (HAR) dataset~\citep{anguita2013public} (openML ID 1478), which consists of smartphone sensor features derived from mechanically coupled signals and is characterized by strong collinearity and non-Gaussianity. After standard pre-processing (Appendix~\ref{app:semireal_preprocess}), we generate random outcome $Y$ by injecting a nonlinear signal via a two-layer MLP,
\[
Y = \mathrm{MLP}_{\theta}(X_S) + \varepsilon,
\]
where all parameters $\theta$ are randomly generated from standard Gaussian,  the support $S$ is fixed across trials and the variance of noise is calibrated to keep an SNR of $0.2$. This setup tests whether the proposed method can recover true signals under a nonlinear model with realistic correlation structures. A robust second-order Gaussian copula is used for knockoff construction (Appendix~\ref{app:gaussian_copula_knockoffs}).

\begin{figure*}[t]
    \centering
    \begin{minipage}{0.32\textwidth}
        \centering
        \includegraphics[width=\linewidth]{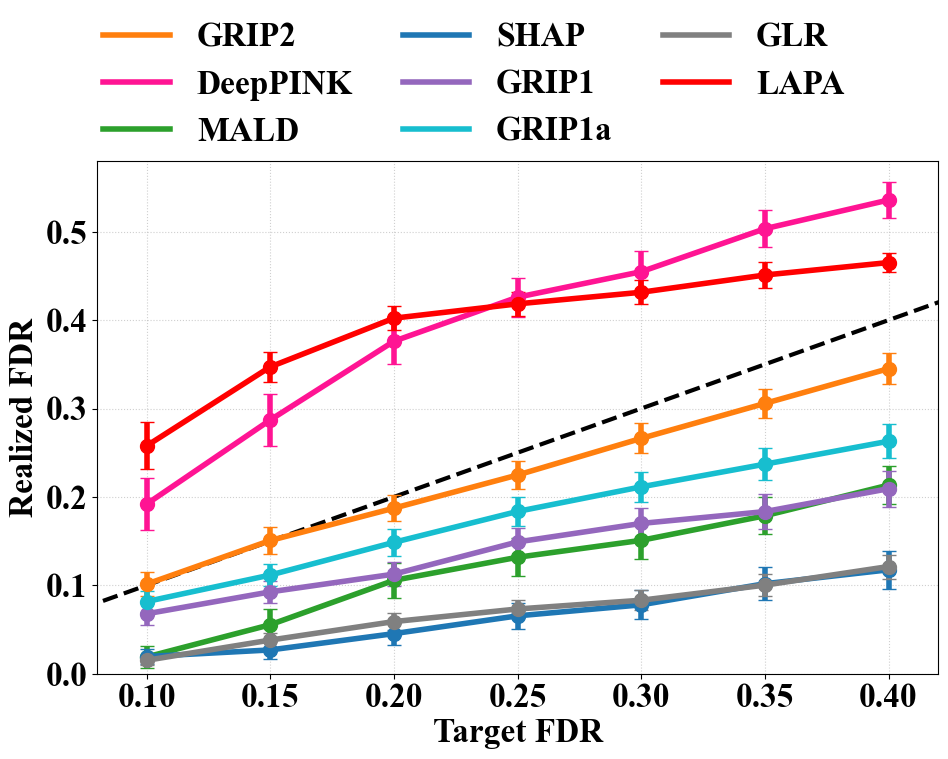}
        \caption{\textbf{FDR validity on HAR data.}
        Realized FDR versus target FDR.
        GRIP2 maintains valid FDR control across all $q$, while DeepPINK and LAPA frequently exceeds the nominal level, indicating invalid FDR control under approximate knockoff construction.}
        \label{fig:har_fdr}
    \end{minipage}
    \hfill
    \begin{minipage}{0.32\textwidth}
        \centering
        \includegraphics[width=\linewidth]{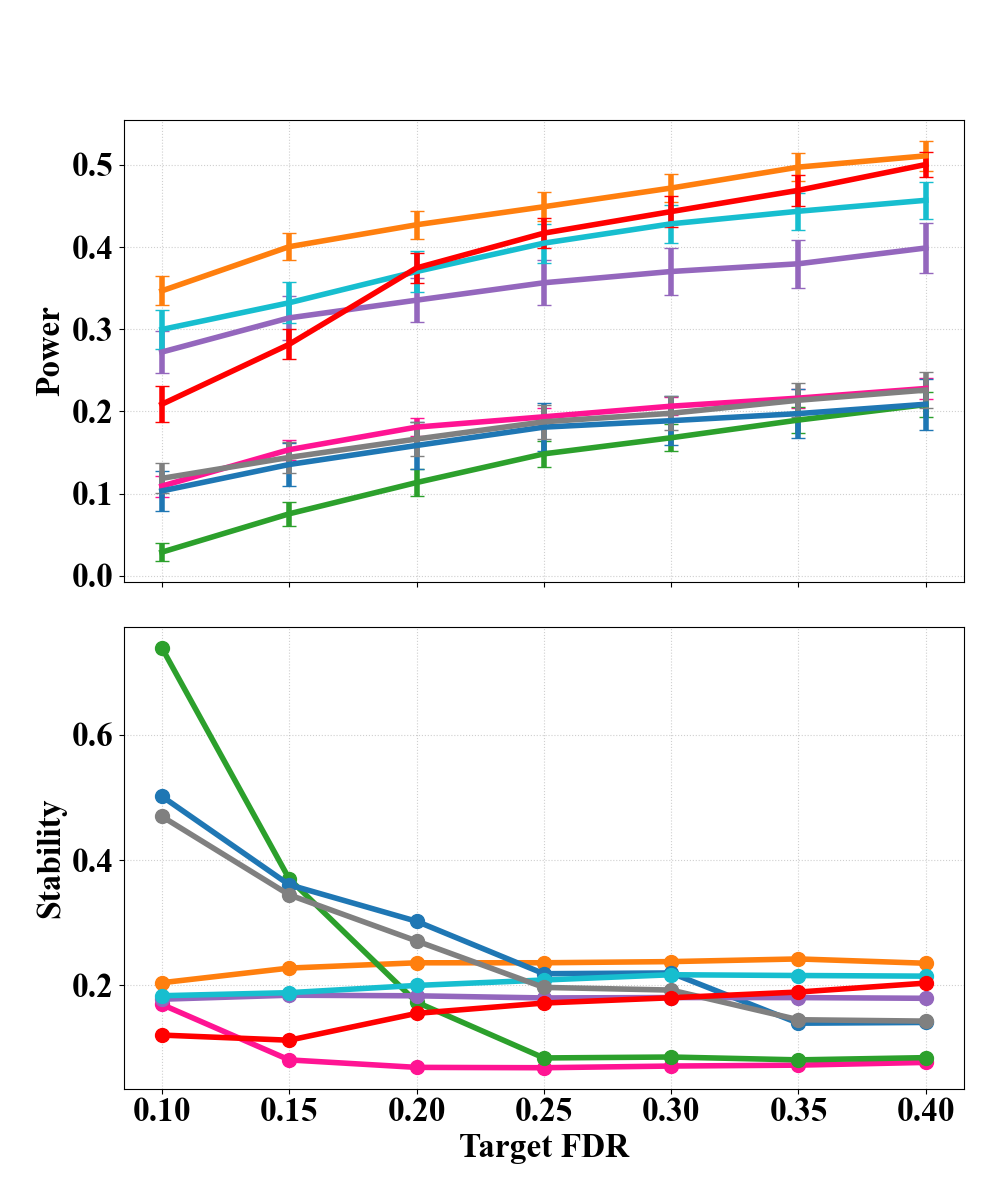}
        \caption{\textbf{Power and Stability vs Target FDR on HAR data.}
        GRIP2 consistently achieve the best power while maintaining favorable stability, whereas other methods either only achieves stability with low power or are worse on both metrics}
        \label{fig:har_power}
    \end{minipage}
    \hfill
    \begin{minipage}{0.32\textwidth}
        \centering
        \includegraphics[width=\linewidth]{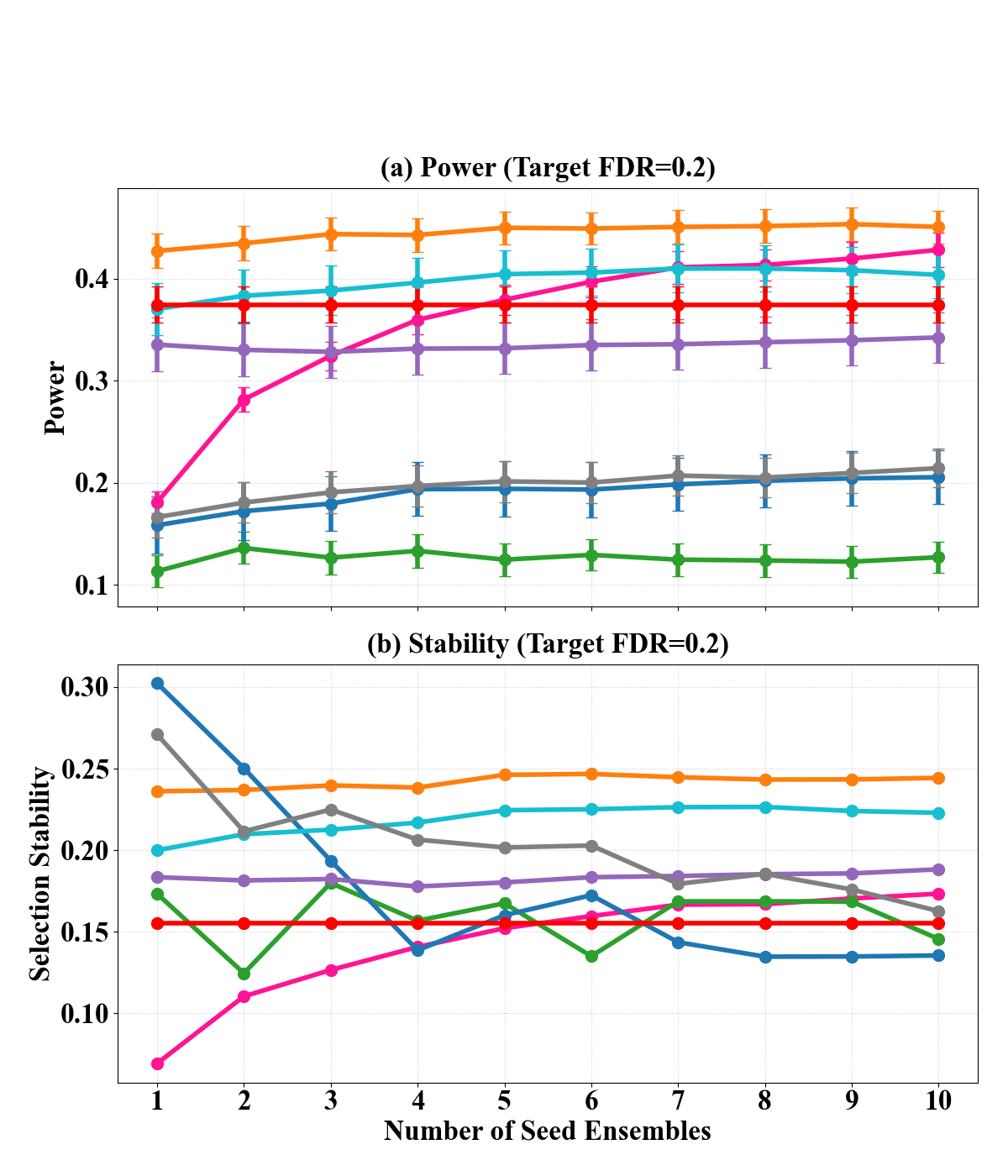}
        \caption{\textbf{Effect of seed ensembling at Target FDR of 20\% on HAR data.}
        Ensembling provides inconsistent benefits for baselines and may reduce stability (e.g., SHAP), whereas GRIP2 achieves stable selections with a single run.}
        \label{fig:har_stability}
    \end{minipage}
\end{figure*}

\textbf{Results.}
Figure~\ref{fig:har_fdr} demonstrates that GRIP2 maintains valid FDR control on the HAR dataset, despite strong collinearity and non-Gaussian distribution. In contrast, DeepPINK and LAPA frequently violates the nominal FDR level, indicating invalid FDR control under approximate knockoff construction. 

Among all methods including those that failed to control FDR, GRIP2 attains the highest power with favorable stability at all target FDRs (Figure~\ref{fig:har_power}). Other methods either perform worse on both metrics consistently or only achieve a better stability with extremely low power. This indicates GRIP2, compared to existing baselines, exhibits an improved ability to recover true signals from a large number of realistic correlated features in a complex nonlinear model.

Consistent with the synthetic experiment, Figure~\ref{fig:har_stability} shows that post-hoc seed ensembling does not reliably improve stability for several baseline methods and can even sometimes degrade it. GRIP2 instead can achieve powerful and stable feature selection with a single training run.

\subsection{Real Data Application: HIV Drug Resistance}
\label{sec:real_data}

\textbf{Dataset and protocol.}
To evaluate performance on real data where the true generative model is unknown, we study the HIV-1 Drug Resistance Database \citep{rhee2003human,rhee2006genotypic, shafer2006rationale}, a standard benchmark for knockoff-based inference \citep{barber2015controlling,candes2018panning,lu2018deeppink}. The features are highly correlated mutations and the goal is to identify resistance-associated mutations. Virology research provides a well-studied reference set of those mutations, which can be used to evaluate the performance of feature selection. 

We focus on Protease Inhibitors (PIs) using binary mutation indicators from HIV-1 protease sequences. Following established practice \citep{barber2015controlling}, we use the \emph{log-fold change} in drug resistance as the response. The association of mutations with this response is routinely modeled using a simple additive linear model, which has a record of strong performance in HIV resistance prediction and knockoff-based analyses~\citep{rhee2006genotypic,vermeiren2007prediction,melikian2012standardized, barber2015controlling}. More details about data pre-processing are reported in Appendix~\ref{app:hiv_preprocess}.

We construct \emph{Fixed-X equi-correlated knockoffs} using the default procedure in the \texttt{knockoff} R package \citep{barber2015controlling}. Fixed-X knockoffs generate synthetic variables that match the Gram structure of the observed design matrix, avoiding distributional assumptions that are difficult to verify for mutation features (details in Appendix~\ref{app:fixedx_knockoffs}).

\textbf{Evaluation.}
We fix the target FDR at 5\%, aligning with the widely used 95\% significance level in biomedical research, and repeat the full procedure over 50 independent trials to assess reproducibility under randomness in both knockoff generation and stochastic optimization (for neural models only). Selection is evaluated against the curated list of Treatment-Selected Mutations (TSMs) \citep{rhee2005hiv}, which is treated as ground truth. We report power, realized FDR (FDP relative to the TSM list), and selection stability measured by average pairwise Jaccard similarity across 50 trials. For a straightforward single-number summary of practical performance, we report metrics averaged over 7 PI drugs; per-drug results are deferred to Appendix~\ref{app:per_drug}.

\textbf{Results.}
Table~\ref{tab:avg_metrics_log} reports the average results based on 50 experiments. GRIP2 controls FDR and achieves the best power and stability.  The lasso path based linear model (LAPA), which attains the best performance in the literature, remains highly competitive in power.
Attribution-based and architecture-driven methods (e.g., SHAP, MALD, DeepPINK) are stable primarily because they make very few ``easy'' discoveries (with DeepPINK unable to select any mutations). Overall, the analysis of HIV data shows that GRIP2 can produce accurate and stable selections on real-world tasks .

\begin{table}[b!]
\centering
\caption{\textbf{HIV-1 Drug Resistance Results.}
Average Power, realized FDR, and selection stability (Jaccard) aggregated over 7 Protease Inhibitor drugs and 50 trials. Target FDR is fixed at $q=0.05$.}
\label{tab:avg_metrics_log}
\resizebox{0.95\linewidth}{!}{%
\begin{tabular}{lcccc}
\toprule
 \textbf{Method} & \textbf{Power (S.E.)} & \textbf{Realized FDR (S.E.)} & \textbf{Stability (S.E.)} \\
\midrule
\textbf{GRIP2} & \textbf{0.222 (0.028)} & \textbf{0.043 (0.009)} & \textbf{0.533 } \\
LAPA & 0.191 (0.027) & 0.037 (0.007) & 0.514  \\
SHAP & 0.070 (0.020) & 0.029 (0.008) & 0.729  \\
MALD & 0.007 (0.006) & 0.006 (0.004) & 0.961  \\
DeepPINK & 0.000 (0.000) & 0.000 (0.000) & 1.000  \\

\bottomrule
\end{tabular}%
}
\end{table}

\section{Discussion and Limitations}
\label{sec:discussion}

This work addresses the challenge of flexible feature selection with FDR control in high correlation and low SNR ratio settings, where multiple feature subsets can achieve comparable predictive performance and traditional feature importance measures are often sensitive to regularization choices and resulting trained model. By defining a new feature importance measure based on the persistence of first-layer group activity across a two-dimensional regularization surface, GRIP2 provides a robust summary of feature importance while remaining compatible with the general Model-X knockoff framework. Numerical experiments demonstrate that the proposed method improves both stability and power relative to those based on existing deep knockoff statistics, for which the simple seed ensembling typically only offers limited improvement. These results suggest that averaging across regularization regimes during training is a useful approach to generate feature importance measures for feature selection.

Several limitations remain. First, GRIP2 does not resolve the fundamental non-identifiability induced by strong correlation: when features are nearly interchangeable with respect to the response, persistence scores should be interpreted as measures of robustness rather than definitive evidence of causal relevance. Second, our implementation focuses on first-layer group sparsification in feedforward networks; extending persistence-based importance to other architectures or structured feature representations is an important direction for future work. Finally, as with all knockoff-based methods, the practical applicability of GRIP2 depends on the availability of high-quality knockoff constructions, which may be challenging in real-world settings where covariate distribution is complex and unknown.

\section*{Impact Statement}

This paper presents work whose goal is to advance the field of Machine
Learning. There are many potential societal consequences of our work, none
which we feel must be specifically highlighted here.

\bibliography{ref}
\bibliographystyle{icml2026}

\newpage
\appendix
\onecolumn
\section{Additional Results}

\subsection{FDR Control Validation (Synthetic Data)}
\label{app:fdr_plots}

In Section~\ref{sec:experiments}, we summarized the False Discovery Rate (FDR) control performance of the evaluated methods. Here, we provide the complete empirical validation curves. Since our synthetic data generation process allows for the construction of exact Model-X knockoffs, all methods equipped with valid anti-symmetric statistics are theoretically guaranteed to control FDR.

\begin{figure}[h!]
    \centering
    \includegraphics[width=0.5\linewidth]{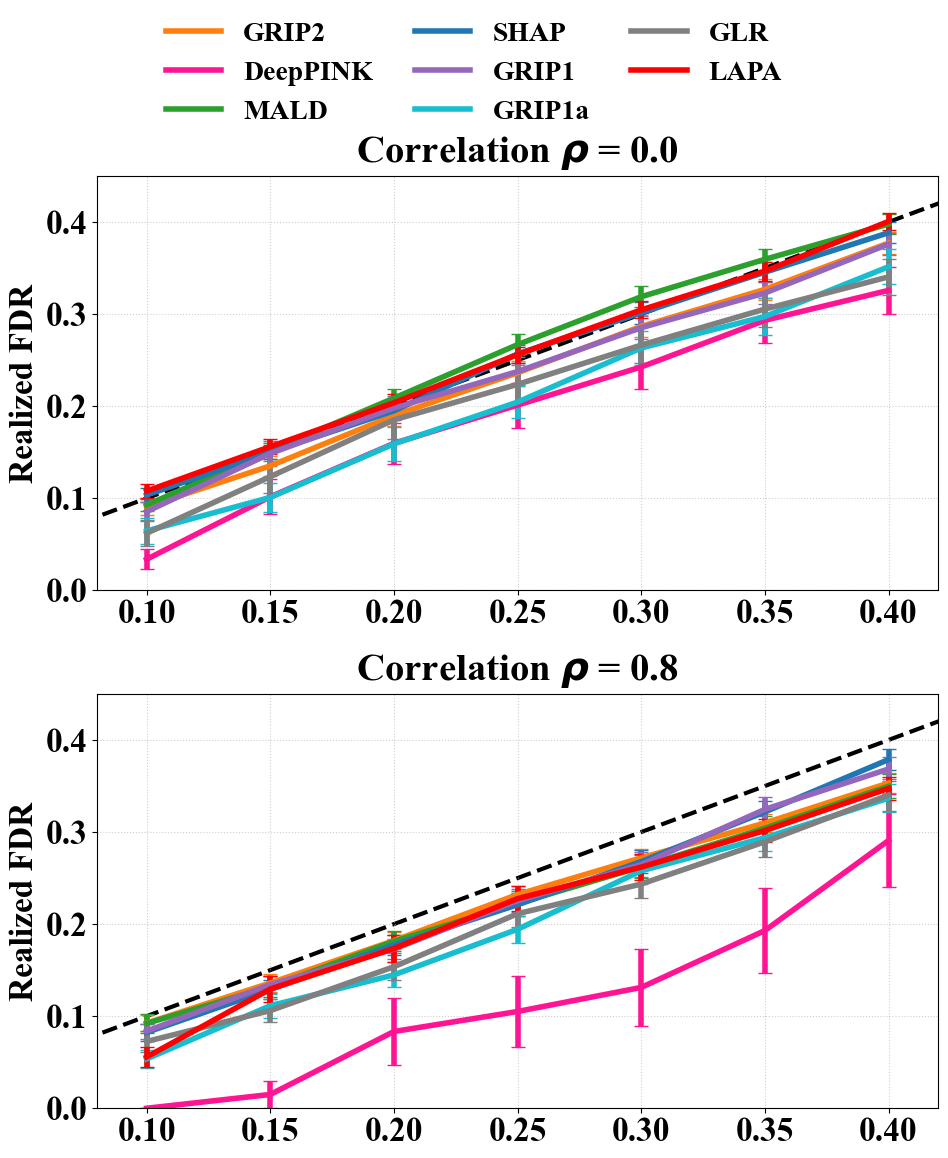}
    \caption{\textbf{FDR Control Validation.} Realized FDR vs. Target FDR ($q$) for varying correlation strengths. The black dashed line ($y=x$) indicates the nominal control level. GRIP2 (orange) strictly respects the FDR limit in all settings.}
    \label{fig:app_fdr_control}
\end{figure}

Figure~\ref{fig:app_fdr_control} confirms this guarantee empirically. We observe:
\begin{enumerate}
    \item \textbf{Global Validity:} Our method, \textbf{GRIP2} (orange), consistently maintains the realized FDR at or below the nominal target level $q$ (dashed line $y=x$) across all correlation regimes ($\rho \in \{0.0, 0.4, 0.8\}$).
    \item \textbf{Baseline Behavior:} Most baselines also control FDR. However, the apparent "control" of DeepPINK (pink) at $\rho=0.8$ is degenerate: as shown in the power analysis (Figure~\ref{fig:synth_rho08}), it achieves low FDR simply by failing to select any features (zero power), rather than by accurately discriminating between signals and nulls.
\end{enumerate}

\subsection{Per-Drug Results on Real-world HIV Drug Resistance Experiments}
\label{app:per_drug}

In Table~\ref{tab:apv} to Table~\ref{tab:nfv}, we present the per-drug performance of each method, averaged over 50 trials. We can see that GRIP2 has significantly better power than other methods in 4 out of 7 drugs, ties first with LAPA for 1 drug and is 2nd best (worse than LAPA) for the rest 2 drugs.

\begin{table}[h!]
\centering
\caption{Performance Metrics for Drug: \textbf{APV}}
\label{tab:apv}
\resizebox{0.8\textwidth}{!}{%
\begin{tabular}{lcccc}
\toprule
\textbf{Target FDR} & \textbf{Method} & \textbf{Power (SE)} & \textbf{Realized FDR (SE)} & \textbf{Jaccard Stability} \\
\midrule

0.05 & GRIP2 & 0.224 (0.032) & 0.026 (0.008) & 0.419 \\
0.05 & LAPA & 0.159 (0.031) & 0.007 (0.004) & 0.494 \\
0.05 & MALD & 0.021 (0.012) & 0.017 (0.010) & 0.884 \\
0.05 & SHAP & 0.028 (0.014) & 0.014 (0.007) & 0.848 \\
0.05 & DeepPINK & 0.000 & 0.000 & 1.000 \\
\bottomrule
\end{tabular}%
}
\end{table}


\begin{table}[h!]
\centering
\caption{Performance Metrics for Drug: \textbf{ATV}}
\label{tab:atv}
\resizebox{0.8\textwidth}{!}{%
\begin{tabular}{lcccc}
\toprule
\textbf{Target FDR} & \textbf{Method} & \textbf{Power (SE)} & \textbf{Realized FDR (SE)} & \textbf{Jaccard Stability} \\
\midrule

0.05 & GRIP2 & 0.058 (0.022) & 0.013 (0.006) & 0.781 \\
0.05 & LAPA & 0.000 & 0.000 & 1.000 \\
0.05 & MALD & 0.000 & 0.000 & 1.000 \\
0.05 & SHAP & 0.019 (0.014) & 0.002 (0.002) & 0.921 \\
0.05 & DeepPINK & 0.000 & 0.000 & 1.000 \\
\bottomrule
\end{tabular}%
}
\end{table}


\begin{table}[h!]
\centering
\caption{Performance Metrics for Drug: \textbf{IDV}}
\label{tab:idv}
\resizebox{0.8\textwidth}{!}{%
\begin{tabular}{lcccc}
\toprule
\textbf{Target FDR} & \textbf{Method} & \textbf{Power (SE)} & \textbf{Realized FDR (SE)} & \textbf{Jaccard Stability} \\
\midrule

0.05 & GRIP2 & 0.368 (0.027) & 0.079 (0.012) & 0.527 \\
0.05 & LAPA & 0.219 (0.031) & 0.062 (0.011) & 0.411 \\
0.05 & MALD & 0.008 (0.008) & 0.006 (0.006) & 0.960 \\
0.05 & SHAP & 0.103 (0.025) & 0.036 (0.009) & 0.593 \\
0.05 & DeepPINK & 0.000 & 0.000 & 1.000 \\
\bottomrule
\end{tabular}%
}
\end{table}


\begin{table}[h!]
\centering
\caption{Performance Metrics for Drug: \textbf{LPV}}
\label{tab:lpv}
\resizebox{0.8\textwidth}{!}{%
\begin{tabular}{lcccc}
\toprule
\textbf{Target FDR} & \textbf{Method} & \textbf{Power (SE)} & \textbf{Realized FDR (SE)} & \textbf{Jaccard Stability} \\
\midrule

0.05 & GRIP2 & 0.106 (0.026) & 0.009 (0.004) & 0.590 \\
0.05 & LAPA & 0.264 (0.033) & 0.037 (0.008) & 0.423 \\
0.05 & MALD & 0.000 & 0.000 & 1.000 \\
0.05 & SHAP & 0.009 (0.009) & 0.004 (0.004) & 0.960 \\
0.05 & DeepPINK & 0.000 & 0.000 & 1.000 \\
\bottomrule
\end{tabular}%
}
\end{table}


\begin{table}[h!]
\centering
\caption{Performance Metrics for Drug: \textbf{NFV}}
\label{tab:nfv}
\resizebox{0.8\textwidth}{!}{%
\begin{tabular}{lcccc}
\toprule
\textbf{Target FDR} & \textbf{Method} & \textbf{Power (SE)} & \textbf{Realized FDR (SE)} & \textbf{Jaccard Stability} \\
\midrule

0.05 & GRIP2 & 0.396 (0.029) & 0.059 (0.010) & 0.535 \\
0.05 & LAPA & 0.209 (0.034) & 0.023 (0.006) & 0.454 \\
0.05 & MALD & 0.012 (0.012) & 0.005 (0.005) & 0.960 \\
0.05 & SHAP & 0.142 (0.028) & 0.054 (0.011) & 0.520 \\
0.05 & DeepPINK & 0.000 & 0.000 & 1.000 \\
\bottomrule
\end{tabular}%
}
\end{table}


\begin{table}[h!]
\centering
\caption{Performance Metrics for Drug: \textbf{RTV}}
\label{tab:rtv}
\resizebox{0.8\textwidth}{!}{%
\begin{tabular}{lcccc}
\toprule
\textbf{Target FDR} & \textbf{Method} & \textbf{Power (SE)} & \textbf{Realized FDR (SE)} & \textbf{Jaccard Stability} \\
\midrule

0.05 & GRIP2 & 0.171 (0.030) & 0.059 (0.012) & 0.457 \\
0.05 & LAPA & 0.255 (0.029) & 0.083 (0.010) & 0.401 \\
0.05 & MALD & 0.000 & 0.000 & 1.000 \\
0.05 & SHAP & 0.112 (0.027) & 0.053 (0.013) & 0.593 \\
0.05 & DeepPINK & 0.000 & 0.000 & 1.000 \\
\bottomrule
\end{tabular}%
}
\end{table}


\begin{table}[h!]
\centering
\caption{Performance Metrics for Drug: \textbf{SQV}}
\label{tab:sqv}
\resizebox{0.8\textwidth}{!}{%
\begin{tabular}{lcccc}
\toprule
\textbf{Target FDR} & \textbf{Method} & \textbf{Power (SE)} & \textbf{Realized FDR (SE)} & \textbf{Jaccard Stability} \\
\midrule

0.05 & GRIP2 & 0.234 (0.032) & 0.056 (0.013) & 0.422 \\
0.05 & LAPA & 0.234 (0.034) & 0.044 (0.009) & 0.417 \\
0.05 & MALD & 0.011 (0.008) & 0.012 (0.008) & 0.921 \\
0.05 & SHAP & 0.075 (0.021) & 0.041 (0.012) & 0.665 \\
0.05 & DeepPINK & 0.000 & 0.000 & 1.000 \\
\bottomrule
\end{tabular}%
}
\end{table}

\section{Knockoff Construction}

\subsection{Second-Order Gaussian Knockoff Construction}
\label{app:gaussian_knockoffs}

This appendix describes the standard second-order Gaussian knockoff construction used when the covariates follow a multivariate normal distribution,
\[
X \sim \mathcal{N}(0, \Sigma),
\]
with known (or well-estimated) covariance matrix $\Sigma \in \mathbb{R}^{p\times p}$.

\paragraph{Goal: second-order matching.}
The Gaussian knockoff framework constructs $\tilde{X}\in\mathbb{R}^{n\times p}$ such that the joint distribution of $(X,\tilde{X})$ is multivariate normal and satisfies the defining second-order conditions
\[
\mathrm{Cov}(\tilde{X}) = \Sigma,
\qquad
\mathrm{Cov}(X,\tilde{X}) = \Sigma - S,
\]
where $S=\mathrm{diag}(s)$ is a nonnegative diagonal matrix chosen to ensure feasibility:
\[
2\Sigma - S \succeq 0.
\]
These conditions imply pairwise exchangeability of $(X_j,\tilde{X}_j)$ under the Gaussian model and yield valid FDR control when combined with an antisymmetric importance statistic.

\paragraph{Equi-Correlated Choice of $S$}

A common and computationally efficient choice is the \emph{equi-correlated} construction \citep{candes2018panning}, which sets all diagonal entries equal: $s_j = s$. In our implementation,
\[
s = \min\{2\lambda_{\min}(\Sigma),\, 1\},
\qquad
s_j = s \ \ \forall j,
\]
where $\lambda_{\min}(\Sigma)$ is the minimum eigenvalue. The cap at $1$ is appropriate when $\Sigma$ is a correlation matrix; more generally, one may cap by $\max_j \Sigma_{jj}$.

\paragraph{Sampling $\tilde{X}$ in Closed Form}

Given $S=\mathrm{diag}(s)$, define
\[
A = \Sigma^{-1} S,
\qquad
M = 2S - S \Sigma^{-1} S.
\]
In practice, we compute $A$ via a linear solve (rather than forming $\Sigma^{-1}$ explicitly), i.e.,
\[
A = \mathrm{Solve}(\Sigma, S).
\]
We then factorize $M$ as
\[
M = C C^\top,
\]
using a Cholesky decomposition. When $\Sigma$ is ill-conditioned or $s$ is near the feasibility boundary, numerical jitter may be required to ensure positive definiteness of $M$.

Finally, we sample an independent Gaussian noise matrix $U \in \mathbb{R}^{n\times p}$ with i.i.d.\ entries $U_{ij}\sim\mathcal{N}(0,1)$ and construct knockoffs as
\[
\tilde{X} \;=\; X(I - A) \;+\; U C^\top.
\]
One can verify that if $X\sim \mathcal{N}(0,\Sigma)$, then $(X,\tilde{X})$ is jointly Gaussian with
\[
\mathrm{Cov}(\tilde{X}) = \Sigma,
\qquad
\mathrm{Cov}(X,\tilde{X}) = \Sigma - S,
\]
as desired.

\paragraph{Numerical safeguards.}
Our implementation symmetrizes $M$ via $(M+M^\top)/2$ and applies a small diagonal jitter (increasing geometrically if needed) before Cholesky factorization to avoid failures due to finite-precision error. We also add a small ridge term when solving linear systems involving $\Sigma$ to improve stability in nearly singular settings.

\paragraph{Returned metadata.}
For reproducibility, we record the covariance $\Sigma$ used for construction and the chosen diagonal $s$ (i.e., the entries of $S$), which fully specify the second-order Gaussian knockoff model used in the experiment.

\subsection{Gaussian Copula Second-Order Knockoff Construction}
\label{app:gaussian_copula_knockoffs}

This appendix details our Gaussian copula based second-order knockoff construction, used for datasets with non-Gaussian and potentially discrete marginals. The implementation follows a standard \emph{copula Gaussianization} pipeline: (i) map each feature to an approximately standard normal variable via rank-based transforms, (ii) fit a covariance model in the Gaussianized space, (iii) generate Gaussian second-order knockoffs, and (iv) map knockoffs back to the original feature marginals via empirical quantiles.

\paragraph{Goal: second-order knockoffs.}
Given an observed design matrix $X \in \mathbb{R}^{n \times p}$ with columns $(X_1,\dots,X_p)$, second-order (approximate) Model-X knockoffs aim to construct $\tilde{X}$ such that the first two moments match those of $X$ and the joint correlation structure between $(X,\tilde{X})$ satisfies the canonical second-order conditions:
\[
\mathrm{Cov}(\tilde{X}) \approx \Sigma,\qquad 
\mathrm{Cov}(X,\tilde{X}) \approx \Sigma - S,
\]
where $\Sigma$ is an estimate of $\mathrm{Cov}(X)$ and $S=\mathrm{diag}(s)$ is a nonnegative diagonal matrix chosen so that
\[
2\Sigma - S \succeq 0.
\]
When combined with an antisymmetric importance statistic, these conditions yield approximate FDR control in practice.

\paragraph{Gaussian Copula Transform and Inverse}

\paragraph{Rank-to-normal transform.}
Because real-world features are often non-Gaussian (e.g., binary indicators, heavy-tailed measurements), we first apply a feature-wise Gaussian copula transform. For each feature $j \in \{1,\dots,p\}$, we compute randomized ranks to obtain pseudo-uniform variables,
\[
U_{ij} \;=\; \frac{r_{ij}+1/2}{n}, \qquad U_{ij}\in(0,1),
\]
where $r_{ij}$ is the rank of $X_{ij}$ among $\{X_{1j},\dots,X_{nj}\}$. To avoid ties (common for discrete features), we use randomized tie-breaking by adding an infinitesimal Gaussian jitter before sorting. We then map $U_{ij}$ to an approximately standard normal variable using the probit transform
\[
Z_{ij} \;=\; \Phi^{-1}(U_{ij}),
\]
where $\Phi$ denotes the standard normal CDF. Numerically, we clip $U_{ij}$ to $[\varepsilon,1-\varepsilon]$ (e.g., $\varepsilon=10^{-6}$) to avoid infinities.

\paragraph{Inverse: empirical quantile mapping.}
After generating knockoffs $\tilde{Z}$ in the Gaussianized space, we map each column $\tilde{Z}_j$ back to the original feature marginal distribution using empirical quantiles:
\[
\tilde{U}_{ij}=\Phi(\tilde{Z}_{ij}), \qquad \tilde{X}_{ij}=F_j^{-1}(\tilde{U}_{ij}),
\]
where $F_j^{-1}$ is approximated by linear interpolation on the empirical CDF grid built from the sorted observed values of feature $j$. Concretely, we precompute the sorted values $\{X_{(1)j}\le \cdots \le X_{(n)j}\}$ and the grid points $u_k=(k+1/2)/n$, and set
\[
\tilde{X}_{ij} \approx \mathrm{Interp}\!\left(\tilde{U}_{ij};\, \{(u_k, X_{(k)j})\}_{k=1}^n\right).
\]
This ensures $\tilde{X}_j$ matches the empirical marginal distribution of $X_j$ by construction.

\paragraph{Covariance Estimation in the Gaussianized Space}

\paragraph{Centering and shrinkage.}
Let $Z \in \mathbb{R}^{n\times p}$ denote the Gaussianized matrix. We first center each column:
\[
Z_c = Z - \mathbf{1}\mu^\top,\quad \mu = \frac{1}{n}\sum_{i=1}^n Z_{i\cdot}.
\]
We then estimate $\Sigma = \mathrm{Cov}(Z)$ using Ledoit--Wolf shrinkage, followed by an additional shrinkage step toward a scaled identity to improve numerical stability in highly collinear settings. Specifically, let $S_{\text{sample}} = \frac{1}{n-1} Z_c^\top Z_c$ and let the shrinkage target be $\mu I$ with $\mu=\mathrm{tr}(S_{\text{sample}})/p$. We form
\[
\widehat{\Sigma}
= (1-\alpha')\, S_{\text{sample}} + \alpha' \, \mu I,
\]
where $\alpha'$ is an ``aggressive'' shrinkage coefficient obtained by adding a small extra shrinkage to the Ledoit--Wolf coefficient and clipping to $[0,1]$. Finally, we symmetrize $\widehat{\Sigma}$ to avoid numerical asymmetry.

\paragraph{Gaussian Second-Order Knockoffs in $Z$-Space}
\label{app:gaussian_second_order}

\paragraph{Choosing $S=\mathrm{diag}(s)$.}
Given the covariance estimate $\widehat{\Sigma}$, we construct a diagonal matrix $S=\mathrm{diag}(s)$ that satisfies $0\preceq S \preceq \widehat{\Sigma}$ and $2\widehat{\Sigma}-S\succeq 0$. In our implementation we use the equi-correlated choice,
\[
s_j = s = \min\{ 2\lambda_{\min}(\widehat{\Sigma}),\ \max_j \widehat{\Sigma}_{jj} \},
\]
where $\lambda_{\min}(\widehat{\Sigma})$ is the smallest eigenvalue of $\widehat{\Sigma}$. This yields a simple, fast, and stable construction.

\paragraph{Constructing $\tilde{Z}$.}
Let $A = \widehat{\Sigma}^{-1} S$, computed via a linear solve rather than explicit inversion. Define
\[
M = 2S - S \widehat{\Sigma}^{-1} S \;=\; 2S - SA.
\]
By construction, $M$ is positive semidefinite (up to estimation error), and we compute a Cholesky factor $C$ such that $CC^\top \approx M$, using a small diagonal jitter if needed.

We then generate Gaussian knockoffs as
\[
\tilde{Z} = Z(\,I - A\,) + U C^\top,
\qquad U \sim \mathcal{N}(0, I_{n\times p}).
\]
One can verify that, conditional on $Z$ being Gaussian with covariance $\widehat{\Sigma}$, the joint covariance of $(Z,\tilde{Z})$ satisfies the second-order knockoff conditions:
\[
\mathrm{Cov}(\tilde{Z}) = \widehat{\Sigma},\qquad
\mathrm{Cov}(Z,\tilde{Z}) = \widehat{\Sigma} - S.
\]

\paragraph{Putting Everything Together}
\label{app:copula_knockoff_pipeline}

The full pipeline is:
\begin{enumerate}
    \item \textbf{Gaussianize marginals:} Apply rank-based Gaussian copula transform to obtain $Z=\Phi^{-1}(U)$ and store per-feature sorted values for inversion.
    \item \textbf{Estimate covariance:} Compute $\widehat{\Sigma}$ from centered $Z$ using Ledoit--Wolf shrinkage plus additional shrinkage toward $\mu I$.
    \item \textbf{Generate knockoffs in Gaussian space:} Construct $S=\mathrm{diag}(s)$ (equi-correlated) and sample $\tilde{Z}=Z(I-A)+UC^\top$ where $A=\widehat{\Sigma}^{-1}S$ and $CC^\top=2S-S\widehat{\Sigma}^{-1}S$.
    \item \textbf{Map back to original marginals:} Set $\tilde{U}=\Phi(\tilde{Z})$ and apply empirical-quantile inversion feature-wise to obtain $\tilde{X}$.
\end{enumerate}

Finally, we build the augmented design matrix used by all knockoff statistics as
\[
X_{2p} = [\,X,\ \tilde{X}\,] \in \mathbb{R}^{n\times 2p}.
\]

\paragraph{Implementation notes.}
We use randomized tie-breaking (infinitesimal jitter) to handle discrete or repeated feature values in the copula transform, clip probabilities away from $\{0,1\}$ to avoid numerical overflow in $\Phi^{-1}$, and apply diagonal jitter during Cholesky factorization when $\widehat{\Sigma}$ is ill-conditioned. These choices improve stability while preserving the intended second-order matching properties.

\subsection{Fixed-X Equi-Correlated Knockoff Construction}
\label{app:fixedx_knockoffs}

We briefly describe the Fixed-X equi-correlated knockoff construction used in all real-data experiments. This procedure matches the default behavior of the \texttt{create.fixed(X, method = "equi")} function in the R \texttt{knockoff} package \citep{barber2015controlling,candes2018panning}.

\paragraph{Problem setup.}
Let $X \in \mathbb{R}^{n \times p}$ denote the observed design matrix, treated as fixed. The goal is to construct a knockoff matrix $\tilde{X} \in \mathbb{R}^{n \times p}$ such that
\[
X^\top X = \tilde{X}^\top \tilde{X}, \quad
X^\top \tilde{X} = X^\top X - \mathrm{diag}(s),
\]
for some vector $s \in \mathbb{R}^p$ satisfying the positive semidefiniteness constraint
\[
2X^\top X - \mathrm{diag}(s) \succeq 0.
\]
These conditions ensure pairwise exchangeability of $(X_j, \tilde{X}_j)$ and guarantee valid FDR control when used with antisymmetric importance statistics.

\paragraph{Equi-correlated choice of $s$.}
In the equi-correlated construction, all coordinates share the same parameter $s_j = s$. Following \citet{barber2015controlling}, we set
\[
s = \min\{1,\, 2 \lambda_{\min}(X^\top X)\},
\]
where $\lambda_{\min}$ denotes the smallest eigenvalue of the Gram matrix. This choice maximizes correlation between original variables and their knockoffs while satisfying the feasibility constraint.

\paragraph{SVD-based construction.}
Let the column-normalized design matrix be written as
\[
X = U D V^\top,
\]
where $U \in \mathbb{R}^{n \times p}$ has orthonormal columns, $D \in \mathbb{R}^{p \times p}$ is diagonal with singular values, and $V \in \mathbb{R}^{p \times p}$ is orthogonal. Let $U_\perp \in \mathbb{R}^{n \times p}$ denote any orthonormal basis for the subspace orthogonal to $U$.

The knockoff matrix is constructed as
\[
\tilde{X}
=
X - X G^{-1} \mathrm{diag}(s)
\;+\;
U_\perp \, \mathrm{diag}\!\left(\sqrt{2s - s^2 / D^2}\right) V^\top,
\]
where $G = X^\top X = V D^2 V^\top$. This construction ensures
\[
\tilde{X}^\top \tilde{X} = G,
\quad
X^\top \tilde{X} = G - \mathrm{diag}(s).
\]

In practice, $U_\perp$ is obtained via a randomized orthogonal completion, which is computationally efficient when $n \gg p$, as in the HIV dataset.

\paragraph{Implementation details.}
We follow the standard implementation used in existing libraries:
(i) columns of $X$ are first normalized to unit $\ell_2$ norm;
(ii) numerical safeguards are applied to clip small negative eigenvalues caused by floating-point error;
(iii) the resulting knockoffs are rescaled back to the original feature norms. The complete implementation is provided in our codebase and closely mirrors the reference R implementation.

\section{Data Pre-processing}

\subsection{Semi-real Benchmark: Data Preprocessing and Target Injection}
\label{app:semireal_preprocess}

This section describes the preprocessing pipeline for our semi-real experiments, which combine real covariates from OpenML with a controlled, synthetic nonlinear signal injection to provide known ground truth. All methods are evaluated on the same processed covariates and injected targets; knockoff construction follows the protocol described in Appendix~\ref{app:copula_knockoff_pipeline}.

\paragraph{OpenML datasets and caching.}
We load tabular datasets from OpenML using the official \texttt{openml} API, with a local cache directory to ensure reproducibility and to avoid system-dependent cache paths. In our semi-real setup we \emph{ignore} any dataset-provided label and treat \emph{all} columns as covariates, since the response is generated by our controlled injection procedure.

\paragraph{Mixed-type tabular preprocessing.}
Given a raw dataframe, we convert it to a dense design matrix $X\in\mathbb{R}^{n\times p}$ via a standard mixed-type pipeline:
\begin{itemize}
    \item \textbf{Numeric columns:} median imputation followed by standardization to zero mean and unit variance.
    \item \textbf{Categorical columns:} most-frequent imputation followed by one-hot encoding (with unseen categories handled gracefully).
\end{itemize}
We record the post-processing feature names and output $X$ in \texttt{float32}. As a safety check (and to match our synthetic setup), we apply an additional global standardization step to the processed matrix:
\[
X \leftarrow \frac{X-\mathrm{mean}(X)}{\mathrm{std}(X)+10^{-8}}
\quad\text{(feature-wise).}
\]

\paragraph{Pre-screening and grouping.}
To reduce redundancy in highly correlated designs---a regime central to this paper---we apply a lightweight, dataset-agnostic pre-screening step prior to knockoff construction. First, we remove near-duplicate features by computing the empirical correlation matrix and dropping any feature whose absolute correlation with an earlier feature exceeds a redundancy threshold of $0.98$. Second, among the remaining features, we form correlation-based clusters using complete-linkage hierarchical clustering on distance
$d(i,j)=1-|\mathrm{corr}(X_i,X_j)|$, and select one representative feature per cluster (default grouping threshold $0.90$). We choose the representative as the feature with the largest empirical variance within the cluster. This step is used only in the semi-real benchmark (not synthetic) to keep the effective dimensionality manageable while preserving the characteristic correlation structure.

\paragraph{Injected nonlinear regression target and ground truth.}
After preprocessing, we generate a semi-real response $y$ by injecting a nonlinear signal on a randomly chosen sparse support. Concretely, we select a support set of size
$s=\lceil 0.2\cdot p\rceil$, extract the supported covariates $X_S$, and define a random two-layer MLP signal
\[
f(X_S)=\big(\sigma(X_S W_1)\big)W_2,
\]
where $\sigma(\cdot)$ is a pointwise nonlinearity (ReLU in our default injector), and $W_1,W_2$ are random weights. We then add Gaussian noise to achieve a prescribed signal-to-noise ratio of $0.2$:
\[
y = f(X_S) + \varepsilon,\qquad 
\varepsilon \sim \mathcal{N}\!\left(0,\ \frac{\mathrm{Var}(f(X_S))}{0.2}\right).
\]
The injected ground truth $S_{\text{true}}\subseteq\{1,\dots,p\}$ is the selected support set used to generate $f(\cdot)$. We repeat injection independently across trials by varying the random seed, while keeping the covariates fixed for each dataset.

\paragraph{Consistency across methods and trials.}
For each dataset and trial, all methods are run on the same $(X,y)$ and evaluated against the same injected support $S_{\text{true}}$. We report power and realized FDR across a grid of target FDR levels, and quantify stability via average pairwise Jaccard similarity of selected feature sets across repeated trials (with the number of trials and the ensemble parameter $K$ specified in the experimental protocol).

\subsection{HIV Data Preprocessing}
\label{app:hiv_preprocess}

This section describes the preprocessing pipeline used in our real-data HIV drug resistance experiments. Knockoff construction for this fixed-design setting is described separately in Appendix~\ref{app:fixedx_knockoffs}.

\paragraph{Raw inputs.}
We use cleaned CSV files containing (i) per-isolate HIV-1 protease sequences encoded by mutation calls at amino-acid positions and (ii) phenotypic drug resistance measurements for multiple drugs within a drug class (e.g., Protease Inhibitors). In addition, we use a curated list of Treatment-Selected Mutations (TSMs) that provides position-level reference annotations for evaluation.

\paragraph{Design matrix construction.}
Starting from per-position mutation strings, we build a sparse binary design matrix of mutation indicators. For each isolate and each sequence position column (e.g., \texttt{P1}, \texttt{P2}, \dots), we parse the reported amino-acid characters and create one binary feature per observed mutation:
\[
\texttt{P}k.\texttt{A} \;\; \text{indicates amino acid A at position } k.
\]
We then pivot these records into an $n\times p$ matrix $X\in\{0,1\}^{n\times p}$ with rows indexed by isolates and columns indexed by mutation indicators, filling missing entries with $0$. We drop any all-zero columns (mutations never observed in the cohort).

\paragraph{Per-drug response extraction and transformation.}
For each drug, we extract the corresponding resistance phenotype vector and discard isolates with missing measurements for that drug. Following standard practice in this benchmark, we use the log-fold change in resistance; concretely, for non-missing measurements $y_{\text{raw}}$, we set
\[
y \;=\; \log\!\left(y_{\text{raw}}\right).
\]
To remove scale differences across drugs and ensure comparable optimization dynamics across trials, we standardize the response within each drug:
\[
y \leftarrow \frac{y-\bar y}{\mathrm{sd}(y)+10^{-8}}.
\]

\paragraph{Feature filtering and duplicate removal.}
To reduce extreme sparsity and improve numerical stability, we apply two simple filters on the mutation indicators for each drug-specific cohort:
(i) we remove features observed in fewer than three isolates (column sum $<3$), and
(ii) we remove duplicate columns (exactly identical mutation indicators), retaining a single representative. After filtering, we obtain the final per-drug design matrix $X\in\mathbb{R}^{n\times p}$ used by all methods.

\paragraph{Ground-truth mapping for evaluation.}
Evaluation is performed at the \emph{position} level using the curated TSM list. We map each mutation feature name (e.g., \texttt{P82.V}) to its numeric position (here $82$) via a regular expression and mark a feature as ``true'' if its position appears in the TSM list for the drug class. During evaluation, a trial's selected set is converted to a set of selected positions; power and FDP are then computed with respect to the TSM position set, and stability is measured by average pairwise Jaccard similarity of selected-position sets across repeated trials.

\paragraph{Training inputs.}
For each drug, we pass the processed $(X,y)$ to the knockoff pipeline and construct fixed-$X$ equi-correlated knockoffs (Appendix~\ref{app:fixedx_knockoffs}). All nonlinear methods then operate on the augmented design $[X,\tilde X]\in\mathbb{R}^{n\times 2p}$ with the same preprocessing and trial protocol.

\section{Implementation Details}
\label{app:impl}

This appendix summarizes implementation details for the proposed GRIP methods and all baselines. Unless otherwise stated, each method operates on the knockoff-augmented design matrix $[X,\tilde X]\in\mathbb{R}^{n\times 2p}$ and produces an importance vector $S\in\mathbb{R}^{2p}$. We then form the knockoff statistic
\[
W_j \;=\; S_j - S_{j+p}, \qquad j=1,\dots,p,
\]
and apply the Model-X knockoff filter to obtain selections with finite-sample FDR control \citep{candes2018panning}.

All neural methods are trained with Adam on a fixed training budget. Randomness from data shuffling, initialization, and any stochastic components is controlled via fixed seeds for reproducibility. For ensemble baselines, we equalize compute by fixing the \emph{total} number of optimization steps and distributing them across ensemble members.

\subsection{Training Configuration}
\label{app:train_cfg}

We use different training configurations depending on the experimental setting. Across settings, neural predictors are optimized with learning rate $10^{-3}$. To decouple sparsity induction from function fitting, we apply structured sparsity only to the \emph{first layer} and apply a separate $\ell_2$ penalty only to \emph{deeper} layers (excluding the first layer).

\paragraph{Synthetic data.}
We use a moderately expressive multilayer perceptron (MLP) with depth $3$ and hidden width $512$, trained for $5{,}000$ optimization steps. GRIP samples regularization parameters from
\[
\log_{10}(\lambda) \sim  \mathcal{U}[-4,-1],\qquad a\sim  \mathcal{U}[a_{\min},1]\;\;\text{with }a_{\min}=0.1,
\]
using block size $M=25$.

\paragraph{Semi-real data.}
We increase the training budget to $10{,}000$ steps. The regularization strength range is shifted to $\lambda\in[1,100]$ to match the larger magnitude and heterogeneity typical of tabular real covariates with injected targets. The deeper-layer $\ell_2$ coefficient is set to $10^{-7}$. Block size is set to $M=25$. Other settings are kept the same.

\paragraph{Real data.}
For the HIV experiment, we use a lightweight MLP (depth $1$, hidden width $1$) with batch size equal to the number of samples to reduce variance and overfitting. The deeper-layer $\ell_2$ coefficient is set to $10^{-2}$, and we use a narrow regularization range $\lambda\in[10^{-3},4\times 10^{-2}]$. Block size is set to $M=50$.

\subsection{Persistence Training Procedure (GRIP)}
\label{app:impl_persistence}

GRIP is implemented by training a single predictor while stochastically varying structured regularization and averaging first-layer activity over the resulting sequence of regularization regimes. This subsection specifies (i) the objective, (ii) the regularization sampling schedules, and (iii) how persistence scores are computed.

\paragraph{Model and grouping.}
Let $m=2p$ denote the augmented input dimension. We train a predictor $g_\theta:\mathbb{R}^m\to\mathbb{R}$ for regression (or to $\mathbb{R}$ with a logistic loss for binary classification). The first affine layer has weights $W^{(0)}\in\mathbb{R}^{h\times m}$. We define the group associated with coordinate $j\in[m]$ as the column vector $w_j := W^{(0)}_{:,j}\in\mathbb{R}^h$.

\paragraph{Training objective.}
Each optimization step minimizes a supervised loss plus two regularizers:
\begin{equation}
\label{eq:grip_obj}
\min_{\theta}\;\;
\mathcal{L}\!\left(g_\theta(x),y\right)
\;+\;
\lambda \,\Omega_a\!\left(\{w_j\}_{j=1}^m\right)
\;+\;
\frac{\gamma}{2}\sum_{\ell\ge 1}\|W^{(\ell)}\|_F^2,
\end{equation}
where $\mathcal{L}$ is mean squared error for regression and logistic loss for binary tasks. The penalty $\Omega_a(\cdot)$ is a \emph{group quasi-norm} applied \emph{only} to the first-layer groups $\{w_j\}$ and parameterized by a geometry parameter $a$ (with a small smoothing constant $\epsilon>0$ for numerical stability). The coefficient $\gamma\ge 0$ applies an $\ell_2$ penalty to \emph{deeper} layers only, mimicking standard weight decay while ensuring that sparsity is driven by the structured first-layer penalty. We use Adam for optimization and apply global gradient clipping (maximum norm $1.0$) in the more challenging regimes.

\paragraph{Sampling schedules for $(\lambda,a)$.}
GRIP repeatedly evaluates feature activity under different regularization regimes by sampling $(\lambda,a)$ from specified ranges. We consider the following schedules:
\begin{itemize}
    \item \textbf{2D-block (default):} sample $(\lambda,a)$ once per block of $M$ steps and keep them fixed within the block.
    \item \textbf{1D-block:} sample $\lambda$ once per block and fix $a=1$ (group-Lasso geometry).
    \item \textbf{1D-block-$a$:} sample $a$ once per block and fix $\lambda=\lambda_{\mathrm{fix}}$.
    \item \textbf{Fixed (single-shot):} keep $(\lambda,a)$ fixed throughout training (baseline).
\end{itemize}
In schedules that require a fixed strength, we use the geometric mean
\[
\lambda_{\mathrm{fix}} = \sqrt{\lambda_{\min}\lambda_{\max}},
\qquad a_{\mathrm{fix}}=1.
\]

\paragraph{Persistence snapshots and the final score.}
Let $\mathcal{A}_\theta\in\mathbb{R}^m_{\ge 0}$ denote the per-feature first-layer activity vector computed from the groups $\{w_j\}$; throughout, $\mathcal{A}_{\theta,j}$ is proportional to $\|w_j\|_2$ and serves as a direct proxy for whether feature $j$ is active at the first-layer gate. GRIP defines the persistence score as an average of activities across multiple sampled regimes:
\begin{equation}
\label{eq:grip_S}
S \;=\; \frac{1}{B}\sum_{b=1}^{B}\mathcal{A}_{\theta^{(t_b)}} \in \mathbb{R}^{m},
\end{equation}
where $\{\theta^{(t_b)}\}_{b=1}^B$ are the parameter values at snapshot times.
For block schedules, we take one snapshot at each block boundary; for minibatch sampling, we take one snapshot per step; for the fixed baseline, we snapshot every $M$ steps to match the block granularity. Warm-up steps do not contribute snapshots.

\paragraph{Diagnostics.}
To monitor the balance between prediction fitting and sparsity pressure, we optionally track the ratio
\[
\frac{\lambda \|\nabla_{W^{(0)}}\Omega_a\|}{\|\nabla_{W^{(0)}}\mathcal{L}\|+\epsilon},
\]
evaluated on the first-layer weights at periodic intervals, where a small positive constant $\epsilon$ is introduced for numerical stability.

\subsection{GRIP Variants and Ablations}
\label{app:impl_grip_variants}

All GRIP variants share the objective \eqref{eq:grip_obj} and differ only in the sampling schedule for $(\lambda,a)$ and how snapshots are aggregated.

\paragraph{Single-shot group Lasso.}
We fix $a=1$ and $\lambda=\lambda_{\mathrm{fix}}$, train once, and define $S$ from first-layer activity snapshots taken every $M$ steps (equivalently, a time-average at fixed regularization).

\paragraph{Persistence-1D and Persistence-2D.}
Persistence-1D samples $\lambda$ in blocks while fixing $a=1$. Persistence-2D samples both $(\lambda,a)$ in blocks, yielding a surface-averaged importance estimate.

\paragraph{Ensembled persistence (when used).}
To isolate the effect of averaging from increased compute, we use a fixed total training budget and split the optimization steps across $K$ independent runs; the final importance vector is $S=K^{-1}\sum_{k=1}^K S^{(k)}$.

\subsection{Lasso-Based Baselines}
\label{app:impl_lasso}.
\paragraph{Entry time along the Lasso path.}
We compute the Lasso path over a decreasing grid of regularization strengths and define $S_j$ as the largest regularization value at which feature $j$ becomes nonzero (earlier entry implies higher importance).

\subsection{Gradient-Based Importance (MALD)}
\label{app:impl_mald}

The Mean Absolute Local Derivative (MALD) baseline first trains a nonlinear MLP with the same training configuration but \emph{without} sparsity-inducing regularization (i.e., with $\lambda=0$ and $\gamma=0$) and then computes importance using input gradients:
\[
S_j \;=\; \frac{1}{n}\sum_{i=1}^n \left|\frac{\partial g_\theta(x_i)}{\partial x_{ij}}\right|^r,
\]
with $r=1$ in our experiments. Gradients are computed by automatic differentiation and averaged over a subset of batches to control cost. An ensemble variant averages $S$ across $K$ independent training runs under a fixed total budget.

\subsection{DeepPINK}
\label{app:impl_deeppink}

We include DeepPINK \citep{lu2018deeppink}, which introduces a pairwise-connected layer that couples $(X_j,\tilde X_j)$ and promotes feature competition within the architecture. Importance is computed from the DeepPINK scoring rule based on the paired activations/weights. We follow the original DeepPINK design choices and recommended training settings, and refer readers to \citet{lu2018deeppink} for architectural and theoretical details.

\subsection{SHAP}
\label{app:impl_shap}

As a post-hoc attribution baseline, we train a standard MLP with the same configuration but without sparsity-inducing regularization and compute SHAP values using a DeepExplainer-style~\citep{shrikumar2017learning} estimator. We define importance by the expected absolute SHAP value across an evaluation set:
\[
S_j \;=\; \mathbb{E}\!\left[\,|\mathrm{SHAP}_j(x)|\,\right].
\]
To reduce variance, we also consider an ensemble variant that averages $S$ across multiple independent runs under a fixed total training budget.

\end{document}